\documentclass{article} 

\usepackage{ifthen}
\usepackage{graphicx}
\usepackage{amssymb} 
\usepackage{amsmath}
\usepackage{times} 
\usepackage[colorlinks=true,linkcolor=blue]{hyperref}

\newcommand{\bookchapterold}[1]{}
\newcommand{\tutorialold}[1]{}
\newcommand{\tutorial}[1]{{#1}}
\newcommand{\bookchapter}[1]{{#1}}

\newenvironment{einschub}{
\begin{small} \begin{quotation}\noindent}{%
  \vspace{0.5ex}\end{quotation}\end{small}
}
\title{The CMA Evolution Strategy: \tutorial{A
    Tutorial}}
\author{Nikolaus Hansen 
\\
   Inria
\\
  Research centre Saclay--\^Ile-de-France
}
\date{\vspace{-1ex}}
\newcommand{\sstrich}[1]{{-}\linebreak[2]#1}

\newcommand{\tabverb}{\hspace*{0.0ex}\verb}
\newcommand{\myframe}[1]{ 
\vspace{0.5ex}\par\noindent
      \frame{\parbox{\textwidth}{#1}}\vspace{0.2ex}
  }

\newcommand{\equationframe}[1]{\myframe{#1}}

\newcommand{\R}{\mathbb{R}}
\newcommand{\kom}[1]{}
\newlength{\RHSbreite}
\newcommand{\where}[1]{%
  \begin{description}
  \item[]#1 \vspace{-0.9ex}
  \end{description}
}
\newcommand{\whereende}{\vspace{2ex}~\\}

\newcommand{\ve}[1]{\mathchoice{\mbox{\boldmath$\displaystyle#1$}}
  {\mbox{\boldmath$\textstyle#1$}} {\mbox{\boldmath$\scriptstyle#1$}}
  {\mbox{\boldmath$\scriptscriptstyle#1$}}}
\newcommand{\ma}[1]{\mathchoice{\mbox{\boldmath$\displaystyle#1$}}
  {\mbox{\boldmath$\textstyle#1$}} {\mbox{\boldmath$\scriptstyle#1$}}
  {\mbox{\boldmath$\scriptscriptstyle#1$}}}

\newcommand{\T}{\ensuremath{\mathsf{T}}}

\newcommand{\mal}{\cdot}

\newcommand{\auge}[1]{\textbf{#1}}

\newcommand{\fsphere}{f_{\mathrm{sphere}}}
\newcommand{\flinear}{f_{\mathrm{linear}}}

\newcommand{\emphindex}[1]{\emph{#1}\index{#1}}
\newcommand{\klamstri}[2]{(#1\mbox{-)}\linebreak[0]#2}

\newcommand{\N}{\ensuremath{n}}
\newcommand{\Nat}{\mathbb{N}}
\newcommand{\Rn}{\R^{\N}}
\newcommand{\rklam}[1]{\left({#1}\right)}

\newcommand{\cma}{CMA-ES}
\newcommand{\gen}{\ensuremath{g}}
\newcommand{\mueff}{\ensuremath{\mu_\mathrm{eff}}}
\newcommand{\sumw}[1][j]{\ensuremath{{\textstyle\sum{\!w_{#1}}}}}
\newcommand{\mucov}{\mu_\mathrm{cov}}
\newcommand{\alphamuminus}{\ensuremath{\alpha_{\mu}^{-}}}
\newcommand{\summ}[2]{\sum_{#1=1}^{#2}}
\newcommand{\suml}[2]{\sum_{#1=0}^{#2}}
\newcommand{\lam}{\ensuremath{\lambda}}
\newcommand{\sig}{\ensuremath{\sigma}}
\newcommand{\sigg}{\sigma^{(g)}}
\newcommand{\siggg}{\sigma^{(g+1)}}
\newcommand{\Bg}{\ma{B}^{(g)}}
\newcommand{\Cg}{\ensuremath{\ma{C}^{(g)}}}
\newcommand{\Dg}{\ma{D}^{(g)}}
\newcommand{\Cgg}{\ma{C}^{(g+1)}}
\newcommand{\Cii}{\ma{C}^{(i+1)}}
\newcommand{\cc}{\ensuremath{c_\mathrm{c}}}
\newcommand{\pc}{\ve{p}_\mathrm{c}}
\newcommand{\pcg}{{\ve{p}_\mathrm{c}^{(g)}}}
\newcommand{\pcgg}{\ve{p}_\mathrm{c}^{(g+1)}}
\newcommand{\cs}{c_\sig}
\newcommand{\ds}{d_\sig}
\newcommand{\ps}{\ve{p}_\sig}
\newcommand{\psgminone}{{\ve{p}_\sig^{(g-1)}}}
\newcommand{\psg}{{\ve{p}_\sig^{(g)}}}
\newcommand{\psgg}{{\ve{p}_\sig^{(g+1)}}}

\newcommand{\xgg}{\ve{x}^{(g+1)}}

\newcommand{\x}{\ve{x}}
\newcommand{\y}{\ve{y}}
\newcommand{\z}{\ve{z}}
\newcommand{\ygg}{\y^{(g+1)}}
\newcommand{\zgg}{\z^{(g+1)}}
\newcommand{\ilam}{{i:\lam}}
\newcommand{\Normal}[1]{{\mathcal N}\hspace{-0.13em}\left(#1\right)}
\newcommand{\Normali}[2][]{{\mathcal N}_{#1}\hspace{-0.13em}\left(#2\right)}
\newcommand{\Id}{\mathbf{I}}
\newcommand{\NormalNullI}{{\mathcal N}
        \hspace{-0.13em}\left({\ve{0},\Id}\right)}
\newcommand{\textNormal}[1]{\mathcal{N}\hspace{-0.13em}(#1)}
\newcommand{\ccov}{\ensuremath{c_{\mathrm{cov}}}}
\newcommand{\cm}{\ensuremath{c_\mathrm{m}}}%
\newcommand{\cone}{\ensuremath{c_1}}
\newcommand{\cmu}{\ensuremath{c_\mu}}
\newcommand{\chiN}{\mathsf{E}\|\NormalNullI\|}

\renewcommand{\whereende}{\vspace{1ex}~\\}
\renewcommand{\mal}{\times}
\newcommand{\xmean}{\ensuremath{\ve{m}}}
\newcommand{\xmeang}{\ensuremath{\xmean^{(g)}}}
\newcommand{\xmeangg}{\ensuremath{\xmean^{(g+1)}}}
\newcommand{\rmDelta}{\mathrm{\Delta}}
\newcommand{\invsqrtC}[1]{{\ma{C}\ifthenelse{%
        \equal{#1}{}}{}{^{(#1)}}}^{-\frac{1}{2}}} 
\newcommand{\sqrtC}[1]{{\ma{C}\ifthenelse{%
        \equal{#1}{}}{}{^{(#1)}}}^{\frac{1}{2}}} 
\newcommand{\invC}[1]{{\ma{C}^{(#1)}}^{-{1}}} 
\newcommand{\ew}{d^2}
\newcommand{\sqrtew}{d}

\newcommand{\nref}[1]{\nolinebreak[4]\hspace{0.2845em plus0.05em minus0.17em}\nolinebreak[4]\ref{#1}}
\newcommand{\pref}[1]{\mbox{.\hspace{0.235em}\ref{#1}}}
\newcommand{\eqrefadd}[1]{(\ref{#1})} 
\newcommand{\eqsref}[1]{(\ref{#1})}
\newcommand{\eqeqref}[1]{equation \eqref{#1}}
\newcommand{\Eqref}[1]{Equation (\ref{#1})}
\newcommand{\Eqsref}[1]{Equations (\ref{#1})}
\newcommand{\figpart}[1]{(\textbf{#1})}

\newcommand{\Extext}[1]{{\mathsf{E}}\hspace{-0.0em}\big[#1\big]}
\begin{document}
\enlargethispage{5ex}
\maketitle              
~\\[-4.2\baselineskip]
\begin{samepage}
  \tableofcontents
  {\tiny\noindent Compiled \today}
\end{samepage}
\newpage
\newcommand{\altneu}[2]{#2}
\newcommand{\ie}{\textit{i.e.}}
\newcommand{\eg}{\textit{e.g.}}
\newcommand{\gstrich}{\hspace{0.3ex}---\hspace{0.3ex}}
\tutorial{
\section*{Nomenclature}\addcontentsline{toc}{section}{Nomenclature}
We adopt the usual vector notation, where bold letters, $\ve{v}$, are
column vectors, capital bold letters, $\ma{A}$, are matrices, and a
transpose is denoted by $\ve{v}^\T$.  A list of used abbreviations and
symbols is given in alphabetical order.  }

\paragraph{Abbreviations}
\begin{description}\addtolength{\itemsep}{-0.9ex}
\item CMA Covariance Matrix Adaptation
\item EMNA Estimation of Multivariate Normal Algorithm
  \kom{\cite{Larranaga:2002,Larranaga:2001}}
\item ES Evolution Strategy 
\item $(\mu/\mu_\mathrm{\{I,W\}},\lambda)$-ES, Evolution Strategy with
  $\mu$ parents, with recombination of all $\mu$ parents, either
  Intermediate or Weighted, and $\lambda$ offspring.
\item RHS Right Hand Side. 
\end{description}

\newcommand{\possum}[1]{\ensuremath{\sum|#1|^+}}%
\newcommand{\negsum}[1]{\ensuremath{\sum|#1|^-}}%

\paragraph{Greek symbols}
\begin{description}\addtolength{\itemsep}{-0.9ex}
\item $\lam\ge2$, population size, sample size, number of offspring,
    see \eqref{eqmut}. 
\item $\mu\le\lam$ parent number, number of (positively) selected search points
  in the population, number of strictly positive recombination weights, see \eqref{eqreco}. 
\item{$\mueff = \rklam{\summ{i}{\mu}w_i^2}^{-1}$, the variance
  effective selection mass for the mean, see \eqref{eqmueff}. }
\item{$\sumw = \summ{i}{\lam}w_i$, sum of all weights, note that $w_i\le0$ for $i>\mu$, see also \eqsref{eqcov} and \eqrefadd{eq-def-w}. } 
\item{$\possum{w_i} = \summ{i}{\mu}w_i = 1$, sum of all positive weights. } 
\item{$\negsum{w_i} = -(\sumw - \possum{w_i}) = -\sum_{i=\mu+1}^\lam w_i\ge0$, minus the sum of all 
  negative weights. } 
\item{$\sigg\in\R_{>0}$, step-size.}
\end{description}

\paragraph{Latin symbols}
\begin{description}\addtolength{\itemsep}{-0.9ex}
\item{$\ma{B}\in\Rn$, an orthogonal matrix. Columns of $\ma{B}$ are
    eigenvectors of $\ma{C}$ with unit length and correspond to the
    diagonal elements of $\ma{D}$.  }
\item{$\Cg\in\R^{\N\times\N}$, covariance matrix at generation $g$.}
\item $c_{ii}$, diagonal elements of $\ma{C}$.
\item $\cone\le1-\cmu$, learning rate for the rank-one update of 
   the covariance matrix update, see
  \eqrefadd{eqrankone}, \eqrefadd{eqcov}, and
  \eqrefadd{algcov}, and Table\nref{tabdefpara}.
\item $\cmu\le1-\cone$, learning rate for the rank-$\mu$ update of 
   the covariance matrix update, see
  \eqsref{eqrankmu}, \eqrefadd{eqcov}, and
  \eqrefadd{algcov}, and Table\nref{tabdefpara}.
\item $\cs<1$, decay rate for the cumulation path for the step-size
  control, see \eqsref{eqcumsig} and \eqrefadd{algps}, and
  Table\nref{tabdefpara}.
\item $\cc\le1$, decay rate for cumulation path for the rank-one update
  of the covariance matrix, see \eqsref{eqpc} and \eqrefadd{algpc},
  and Table\nref{tabdefpara}.
\item $\cm = 1$, learning rate for the mean.
\item{$\ma{D}\in\Rn$, a diagonal matrix. The diagonal elements of
    $\ma{D}$ are square roots of eigenvalues of $\ma{C}$ and
    correspond to the respective columns of $\ma{B}$. }
\item $\sqrtew_i>0$, diagonal elements of diagonal matrix $\ma{D}$, 
    $\ew_i$ are eigenvalues of $\ma{C}$. 
\item $\ds\approx1$, damping parameter for step-size update, see
  \eqsref{eqlogsig}, \eqrefadd{eqsig}, and \eqrefadd{algsig}.
\item $\mathsf{E}$ Expectation value
\item $f: \Rn\to\R, \ve{x}\mapsto f(\ve{x})$, objective function
  (fitness function) to be minimized.
\item $\fsphere: \Rn\to\R,
  \ve{x}\mapsto\|\ve{x}\|^2=\ve{x}^\T\ve{x}=\summ{i}{\N}x_i^2$.
\item{$g\in\Nat_0$, generation counter, iteration number. }
\item $\Id\in\R^{\N\times\N}$, Identity matrix, unity matrix. 
\item{$\xmeang\in\Rn$, mean value of the search
  distribution at generation $g$. }
\item $\N\in\Nat$, search space dimension, see $f$.
\item $\NormalNullI$, multivariate normal distribution with zero mean
  and unity covariance matrix. A vector distributed according to
  $\NormalNullI$ has independent, $(0,1)$-normally distributed
  components.
\item $\Normal{\xmean,\ma{C}}\sim\xmean+\Normal{\ve{0},\ma{C}}$,
  multivariate normal distribution with mean $\xmean\in\Rn$ and
  covariance matrix $\ma{C}\in\R^{\N\times\N}$. The matrix $\ma{C}$ is
  symmetric and positive definite.
\item{$\R_{>0}$, strictly positive real numbers. }
\item{$\ve{p}\in\Rn$, evolution path, a sequence of successive
    (normalized) steps, the strategy takes over a number of
    generations. }
\item{$w_i$, where $i=1,\dots,\lam$, recombination weights, see \eqref{eqreco} and \eqref{eqrankmu} and \eqref{eq-w-a}--\eqref{eq-def-w}.}
\item{$\xgg_k\in\Rn$, $k$-th offspring/individual from generation
    $g+1$. We also refer to $\xgg$, as search point, or object
    parameters/variables, commonly used synonyms are candidate
    solution, or design variables. }
\item{$\xgg_{i:\lam}$, $i$-th best individual out of
  $\xgg_1,\dots,\xgg_\lam$, see \eqref{eqmut}. The index
  \mbox{$i:\lam$} denotes the index of the $i$-th ranked individual
  and $f(\xgg_{1:\lam})\le f(\xgg_{2:\lam})\le\dots\le
  f(\xgg_{\lam:\lam})$, where $f$ is the objective function to be
  minimized.}
\item $\ygg_k=(\xgg_k - \xmeang) / \sigg$ corresponding to $\x_k = \xmean + \sig\y_k$. 
%
\end{description}

\tutorial{
\setcounter{section}{-1}
\section{Preliminaries}
 This tutorial introduces the CMA Evolution Strategy (ES), where CMA
 stands for Covariance Matrix Adaptation.\footnote{%
 Parts of this material have also been presented in \cite{Hansen:2006}
 and \cite{hansen2010habil}, in the context of 
 \emph{Estimation of Distribution Algorithms} and \emph{Adaptive Encoding}, 
 respectively. An introduction deriving CMA-ES from the 
 information-geometric concept of a natural
 gradient can be found in \cite{hansen2014principled}.} 
 The CMA-ES is a stochastic, 
 or \emph{randomized},
 method for real-parameter (continuous domain) optimization of
 non-linear, non-convex functions (see also Section\nref{sec:blackbox}
 below).\footnote{While CMA variants for \emph{multi-objective}
 optimization and \emph{elitistic} variants have been proposed, this
 tutorial is solely dedicated to single objective optimization and
 non-elitistic truncation selection, also referred to as
 comma-selection.  } We try to motivate and derive the algorithm from
 intuitive concepts and from requirements of non-linear, non-convex
 search in continuous domain.
%
 For a concise
 algorithm description see
 Appendix\nref{sec:algorithm}. A respective Matlab source code is
 given in Appendix\nref{sec:source}.

Before we start to introduce the algorithm in
 Sect\pref{sec:basic}, a few required fundamentals are summed up.

\subsection{Eigendecomposition of a Positive Definite Matrix\label{sec:eigen}}
A symmetric, positive definite matrix, $\ma{C}\in\R^{\N\times\N}$, is
characterized in that for all $\ve{x}\in\Rn\backslash\{\ve{0}\}$ holds
$\ve{x}^\T\ma{C}\ve{x}>0$. The matrix $\ma{C}$ has an orthonormal
basis of eigenvectors, $\ma{B}=[\ve{b}_1,\dots,\ve{b}_\N]$, with
corresponding eigenvalues, $\ew_1,\dots,\ew_\N>0$.

\begin{einschub}%
That means for each $\ve{b}_i$ holds
\begin{equation}\label{eqdefeigenvec}
\ma{C}\ve{b}_i=\ew_i\ve{b}_i  \enspace.
\end{equation}
The important message from \eqref{eqdefeigenvec} is that
\emph{eigenvectors are not rotated} by $\ma{C}$. This feature uniquely
distinguishes eigenvectors. Because we assume the orthogonal eigenvectors to be
of unit length, $\ve{b}_i^\T\ve{b}_j = \delta_{ij} = \left\{
  \begin{array}{ll}
    1 & \mbox{~if~}i=j\\
    0 & \mbox{~otherwise}
  \end{array}\right.$, and $\ma{B}^\T\ma{B} = \Id$
  (obviously this means $\ma{B}^{-1}=\ma{B}^\T$, and it follows
  $\ma{B}\ma{B}^\T = \Id$). An basis of eigenvectors is practical,
  because for any $\ve{v}\in\Rn$ we can find coefficients $\alpha_i$,
  such that $\ve{v} = \sum_i \alpha_i\ve{b}_i$, and then we have
  $\ma{C}\ve{v} = \sum_i\ew_i\alpha_i\ve{b}_i$.
\end{einschub}
The eigendecomposition of $\ma{C}$ obeys
\begin{equation}\label{eqeigen}
  \ma{C} = \ma{B}\ma{D}^2\ma{B}^\T \enspace,
\end{equation}
where
\where{$\ma{B}$ is an orthogonal matrix, $\ma{B}^\T\ma{B}=\ma{B}\ma{B}^\T=\Id$. 
  Columns of $\ma{B}$ form an orthonormal basis of eigenvectors. 
   }
\where{$\ma{D}^2 = \ma{D}\ma{D} =
  \mathrm{diag}(\sqrtew_1,\dots,\sqrtew_\N)^2 =
  \mathrm{diag}(\ew_1,\dots,\ew_\N)$ is a diagonal matrix with
  eigenvalues of $\ma{C}$ as diagonal elements.   }
\where{$\ma{D}=\mathrm{diag}(d_1,\dots,d_\N)$ is a diagonal matrix
  with square roots of eigenvalues of $\ma{C}$ as diagonal elements.
   }
\whereende
The matrix decomposition \eqref{eqeigen} is unique, apart from signs
of columns of $\ma{B}$ and permutations of columns in $\ma{B}$ and
$\ma{D}^2$ respectively, given all eigenvalues are different.\footnote{%
  Given $m$ eigenvalues are equal, any orthonormal basis of their
  $m$-dimensional subspace can be used as column vectors.  For $m>1$
  there are infinitely many such bases. }
%

%
  Given the eigendecomposition \eqref{eqeigen}, the inverse
  $\ma{C}^{-1}$ can be computed via
  \begin{eqnarray*}
    \ma{C}^{-1} &=& \rklam{\ma{B}\ma{D}^2\ma{B}^\T}^{-1}\\
                &=& {\ma{B}^\T}^{-1} \ma{D}^{-2}\ma{B}^{-1}\\
                &=& {\ma{B}}\;\ma{D}^{-2}\ma{B}^{\T}\\
                &=& \ma{B}\;
          \mathrm{diag}\rklam{\frac{1}{\ew_1},\dots,\frac{1}{\ew_\N}}
          \ma{B}^{\T} \enspace.
  \end{eqnarray*}
From \eqref{eqeigen} we naturally define the square root of 
  $\ma{C}$ as
\begin{equation}
  \label{eq:sqrtC}
  \sqrtC{} = \ma{B}\ma{D}\ma{B}^\T
\end{equation}
and therefore 
\begin{eqnarray*}
   \invsqrtC{} &=&\ma{B}\ma{D}^{-1}\ma{B}^\T \\
         &=& \ma{B}\;\mathrm{diag} 
              \rklam{\frac{1}{d_1},\dots,\frac{1}{d_\N}} \ma{B}^\T
\end{eqnarray*}

\subsection{The Multivariate Normal Distribution\label{sec:normal}}
A multivariate normal distribution, $\Normal{\ve{m},\ma{C}}$, has a
unimodal, ``bell-shaped'' density, where the top of the bell (the
modal value) corresponds to the distribution mean, $\ve{m}$. The
distribution $\Normal{\ve{m},\ma{C}}$ is uniquely determined by its
mean $\ve{m}\in\Rn$ and its symmetric and positive definite covariance
matrix $\ma{C}\in\R^{\N\times\N}$.  Covariance (positive definite)
matrices have an appealing {geometrical interpretation}: they can be
uniquely identified with the \klamstri{hyper}{ellipsoid}
$\{\ve{x}\in\Rn\,|\,\ve{x}^\T\ma{C}^{-1}\ve{x}=1\}$, as
shown in Fig\pref{figellipsen}.%
\begin{figure}[tb]
  \begin{center}
    \begin{minipage}{\textwidth}
      \includegraphics[width=0.99\textwidth]{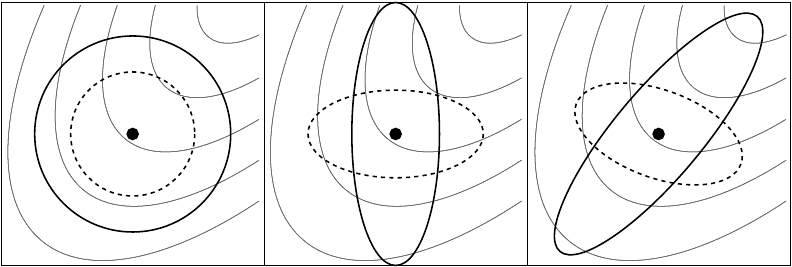}~\\
\parbox{0.33\textwidth}{ 
\centerline{$\Normal{\ve{0},\sigma^2\Id}\kom{\sim\sigma\NormalNullI}$}%
}
\parbox{0.33\textwidth}{ 
\centerline{$\Normal{\ve{0},\ma{D}^2}\kom{\sim\ma{D}\NormalNullI}$}%
}
\parbox{0.30\textwidth}{ 
\centerline{$\Normal{\ve{0},\ma{C}}\kom{\sim\ma{B}\ma{D}\NormalNullI}$}
}\\[-3ex]
    \end{minipage}
  \end{center}
  \caption[Densities]{\label{figellipsen} Ellipsoids depicting 
    one-$\sigma$ lines of equal density of six different normal
    distributions, where $\sigma\in\R_{>0}$, $\ma{D}$ is a diagonal
    matrix, and $\ma{C}$ is a positive definite full covariance
    matrix. Thin lines depict possible objective function contour
    lines }
\end{figure}
The ellipsoid is a {surface} of equal density of the distribution. The
principal axes of the ellipsoid correspond to the eigenvectors of
$\ma{C}$, the squared axes lengths correspond to the eigenvalues.  The
eigendecomposition is denoted by
$\ma{C}=\ma{B}\rklam{\ma{D}}^2\ma{B}^\T$ (see Sect\pref{sec:eigen}).  If
$\ma{D}=\sigma\Id$, where $\sigma\in\R_{>0}$ and $\Id$ denotes the
identity matrix, $\ma{C}=\sigma^2\Id$ and the ellipsoid is isotropic
(Fig\pref{figellipsen}, left). If $\ma{B}=\Id$, then $\ma{C}=\ma{D}^2$
is a diagonal matrix and the ellipsoid is axis parallel oriented
(middle).  In the coordinate system given by the columns of $\ma{B}$, the
distribution $\Normal{\ve{0},\ma{C}}$ is always uncorrelated.

The normal distribution $\Normal{\ve{m}, \ma{C}}$ can be written in
different ways.
\begin{eqnarray}
  \label{eq:normalforms}
\Normal{\ve{m},\ma{C}}&\sim&
\ve{m}+\Normal{\ve{0},\ma{C}}\nonumber\\&\sim&
\ve{m}+\sqrtC{}\NormalNullI \nonumber\\&\sim& 
\ve{m}+\ma{B}\ma{D}\underbrace{\ma{B}^\T\NormalNullI}_{\sim\,\NormalNullI}
            \nonumber\\&\sim&
\ve{m}+\ma{B}\underbrace{\ma{D}\NormalNullI}_{\sim\,\Normal{\ve{0},\ma{D}^2}}  \enspace, 
\end{eqnarray}
where ``$\sim$'' denotes equality in distribution, and
$\sqrtC{}=\ma{B}\ma{D}\ma{B}^\T$. The last row can be well
interpreted, from right to left
\begin{description}
\item[$\NormalNullI$] produces an spherical (isotropic) distribution
  as in Fig\pref{figellipsen}, left.
\item[$\ma{D}$] scales the spherical distribution within the coordinate
  axes as in Fig\pref{figellipsen},
  middle. $\ma{D}\NormalNullI\sim\,\Normal{\ve{0},\ma{D}^2}$ has $n$
  independent components. The matrix $\ma{D}$ can be interpreted as
  (individual) step-size matrix and its diagonal entries are the standard
  deviations of the components.
\item[$\ma{B}$] defines a new orientation for the ellipsoid, where the
  new principal axes of the ellipsoid correspond to the columns of
  $\ma{B}$. Note that $\ma{B}$ has $\frac{n^2-n}{2}$ degrees of freedom. 
\end{description}
\Eqref{eq:normalforms} is useful to compute $\Normal{\ve{m},\ma{C}}$
distributed vectors, because $\NormalNullI$ is a vector of
independent $(0,1)$-normally distributed numbers that can easily be
realized on a computer.
%

\subsection{Randomized Black Box Optimization\label{sec:blackbox}}
We consider the {black box search} scenario, where we want to
\emph{minimize an objective function} (or \emph{cost} function or
\emph{fitness} function)
\begin{eqnarray*} 
f&:& \Rn\to\R \\ && \ve{x}\mapsto f(\ve{x}) \enspace.
\end{eqnarray*}
 The \auge{objective} is to find one or more search points (candidate
 solutions), $\ve{x}\in\Rn$, with a function value, $f(\ve{x})$, as
 small as possible.  We do not state the objective of searching for a
 global optimum, as this is often neither feasible nor relevant in
 practice. \emph{Black box} optimization refers to the situation,
 where function values of evaluated search points are the only
 accessible information on $f$.\footnote{%
   Knowledge about the underlying optimization problem might well enter the
   composition of $f$ and the chosen problem \emph{encoding}. }
 The search points to be evaluated can be freely chosen. We define the
 \auge{search costs} as the number of executed function evaluations,
 in other words the amount of information we needed to acquire from
 $f$\footnote{%
   Also $f$ is sometimes denoted as \emph{cost function}, but it should
   not to be confused with the \emph{search costs}.}. %
  Any performance measure must consider the search costs \emph{together} with
 the achieved objective function value.\footnote{%
  A performance measure can be obtained from a number of trials as,
  for example, the mean number of function evaluations to reach a
  given function value, or the median best function value obtained
  after a given number of function evaluations.}

A randomized black box search algorithm is outlined in
Fig\pref{figbbsearch}.
\begin{figure}[tb]
  \equationframe{
  Initialize distribution parameters $\ve{\theta}^{(0)}$\\
  For generation $g=0,1,2,\dots$\\
  \hspace*{1em}Sample $\lambda$ independent points from
  distribution
  $P\rklam{\ve{x}|\ve{\theta}^{(g)}} \to \ve{x}_1,\dots,\ve{x}_\lam$\\
  \hspace*{1em}Evaluate the sample $\ve{x}_1,\dots,\ve{x}_\lam$ on $f$\\
  \hspace*{1em}Update parameters $\ve{\theta}^{(g+1)} = F_{{\theta}}(\ve{\theta}^{(g)},
    (\ve{x}_1, f(\ve{x}_1)),\dots,(\ve{x}_\lam,f(\ve{x}_\lam)))$\\
  \hspace*{1em}break, if termination criterion met
  }
\caption[Black box search]{\label{figbbsearch}Randomized black box
  search. $f:\Rn\to\R$ is the objective function }
\end{figure}
In the CMA Evolution Strategy the search distribution, $P$, is a
multivariate normal distribution.  Given all variances and covariances, the
normal distribution has the largest entropy of all distributions in
$\Rn$.\kom{http://www.math.uconn.edu/~kconrad/blurbs/entropypost.pdf}
Furthermore, coordinate directions are not distinguished in any way.
Both makes the normal distribution a particularly attractive candidate
for randomized search.
 
 {Randomized search algorithms} are regarded to be robust in a rugged
 search landscape, which can comprise discontinuities, (sharp) ridges, 
 or local optima. The covariance matrix adaptation (CMA) in particular
 is designed to tackle, additionally, ill-conditioned and
non-separable\tutorial{\footnote{%
  An $\N$-dimensional \emph{separable} problem can be solved by
  solving $\N$ $1$-dimensional problems separately, which is a
  far easier task.}}
problems.

} 
%

\subsection{Hessian and Covariance Matrices}
We consider the convex-quadratic objective function
$f_{\ma{H}}:\ve{x}\mapsto \frac{1}{2}\ve{x}^\T\ma{H}\ve{x}$, where the
Hessian matrix $\ma{H}\in\R^{\N\times\N}$ is a positive definite
matrix. Given a search distribution $\Normal{\ve{m},\ma{C}}$, there is
a close relation between $\ma{H}$ and $\ma{C}$: Setting
$\ma{C}=\ma{H}^{-1}$ on $f_{\ma{H}}$ is equivalent to optimizing the
isotropic function
$\fsphere(\ve{x})=\frac{1}{2}\ve{x}^\T\ve{x}=\frac{1}{2}\sum_i x_i^2$
(where $\ma{H}=\Id$) with
$\ma{C}=\Id$.\footnote{%
  Also the initial mean value $\ve{m}$ has to be transformed
  accordingly. }
That is, on convex-quadratic objective functions, setting the
covariance matrix of the search distribution to the inverse Hessian
matrix is equivalent to rescaling the ellipsoid function into a
spherical one.  Consequently, we assume that the optimal covariance
matrix equals to the inverse Hessian matrix, up to a constant 
factor.\footnote{%
  Even though there is good intuition and strong empirical evidence
  for this statement, a rigorous proof is missing. }
Furthermore, choosing a covariance matrix or choosing a respective
affine linear transformation of the search space (\ie\ of $\ve{x}$)
is equivalent \cite{Hansen:2000}, because for any full rank
$\N\times\N$-matrix $\ma{A}$ we find a positive definite Hessian such
that $\frac{1}{2}(\ma{A}\ve{x})^\T \ma{A}\ve{x} = \frac{1}{2}\ve{x}^\T
\ma{A}^\T \ma{A}\ve{x} = \frac{1}{2}\ve{x}^\T\ma{H}\ve{x}$.

The final \auge{objective} of covariance matrix adaptation is
to closely \emph{approximate the contour lines of the objective function $f$}. On convex-quadratic functions this amounts to approximating the inverse Hessian matrix, similar to a quasi-Newton method. 

In
Fig\pref{figellipsen} the solid-line distribution in the right figure
follows the objective function contours most suitably, and it is easy
to foresee that it will aid to approach the optimum the most.

The \label{def:condition}\auge{condition number} of a positive
definite matrix $\ma{A}$ is defined via the Euclidean norm:
$\mathtt{cond}(\ma{A})\stackrel{\mathrm{def}}{=}\|\ma{A}\|\mal\|\ma{A}^{-1}\|$,
where $\|\ma{A}\|=\sup_{\|\ve{x}\|=1}{{\|\ma{A}\ve{x}\|}{}}$. For a
positive definite (Hessian or covariance) matrix $\ma{A}$ holds
$\|\ma{A}\|=\lambda_\mathrm{max}$ and $\mathtt{cond}(\ma{A}) =
\frac{\lambda_\mathrm{max}}{\lambda_\mathrm{min}}\ge1$, where
$\lambda_\mathrm{max}$ and $\lambda_\mathrm{min}$ are the largest and
smallest eigenvalue of $\ma{A}$.
%

\section{Basic Equation: Sampling\label{sec:basic}}
In the CMA Evolution Strategy, a population of new search points
(individuals, offspring) is generated by sampling a multivariate normal
distribution.\footnote{%
   Recall that, given all (co-)variances, the normal distribution has
   the largest entropy of all distributions in $\Rn$. }
The {basic equation} for sampling the search points,
for generation number $g=0,1,2,\dots$, reads\footnote{%
  Framed equations belong to the final algorithm of a CMA Evolution
  Strategy. }
\equationframe{{
\begin{eqnarray}\label{eqmut}
  \ve{x}^{(g+1)}_{k} &\sim& \xmeang + \sigg\Normal{\ve0,\Cg}
  \qquad\mbox{for~}k=1,\dots,\lam 
\end{eqnarray}
}}
where
\where{$\sim$ denotes the same distribution on the left and right side. }
\where{$\textNormal{\ve0,\Cg}$ is a multivariate
  normal distribution with zero mean and covariance matrix \Cg, see
  Sect\pref{sec:normal}. It holds $\xmeang +
  \,\sigg\textNormal{\ve0,\Cg}\sim\Normal{\xmeang,({\sigg})^2\Cg}$.  }
\where{$\ve{x}_k^{(g+1)}\in\Rn$, $k$-th offspring (individual, search point) from
  generation $g+1$.}
%
%
\where{$\xmeang\in\Rn$, mean value of the search
  distribution at generation $g$. }
\where{$\sigg\in\R_{>0}$, ``overall'' standard deviation, step-size, at
  generation $g$.}
\where{$\Cg\in\R^{\N\times\N}$, covariance matrix at generation
  $g$. Up to the scalar factor ${\sigg}^2$, $\Cg$ is the covariance
  matrix of the search distribution. }
\where{$\lam\ge2$, population size, sample size, number
  of offspring. }
\whereende
To define the complete iteration step, the remaining question is, how
to calculate $\xmeangg$, $\Cgg$, and $\siggg$ for the next generation
$g+1$. The next three sections will answer these questions,
respectively. An algorithm summary with all parameter settings and
\textsc{matlab} source code are given in Appendix\nref{sec:algorithm}
and \ref{sec:source}, respectively.

\section{Selection and Recombination: Moving the Mean}

The new mean $\xmeangg$ of the search distribution is a \emph{weighted
  average of $\mu$ selected points} from the sample
$\xgg_1,\dots,\xgg_\lam$\kom{ \cite{Hansen:2001}}\index{weighted
  recombination}:
%
\begin{eqnarray}
  \xmeangg &=& \summ{i}{\mu} w_i\, \ve{x}_{i:\lam}^{(g+1)}
     \label{eqreco}
\end{eqnarray}
\begin{eqnarray}\label{eqweights}
  \sum_{i=1}^{\mu} w_i = 1, \qquad  w_1\ge w_2\ge\dots\ge w_\mu>0
\end{eqnarray}
where 
\where{$\mu\le\lam$ is the parent population size, \ie\ the number of
  selected points. }
\where{$w_{i=1\dots\mu}\in\R_{>0}$, positive weight coefficients for
  recombination.  For $w_{i=1\dots\mu}={1}/{\mu}$, \Eqref{eqreco}
  calculates the mean value of $\mu$ selected points. }
\where{$\xgg_{i:\lam}$, $i$-th best individual out of
  $\xgg_1,\dots,\xgg_\lam$ from \eqref{eqmut}. The index
  \mbox{$i:\lam$} denotes the index of the $i$-th ranked individual
  and $f(\xgg_{1:\lam})\le f(\xgg_{2:\lam})\le\dots\le
  f(\xgg_{\lam:\lam})$, where $f$ is the objective function to be
  minimized.}
\whereende
 \Eqref{eqreco} implements \emph{truncation selection} by choosing
 $\mu<\lam$ out of $\lam$ offspring points.  Assigning
 \emph{different} weights $w_i$ should also be interpreted as a
 selection mechanism. \Eqref{eqreco} implements \emph{weighted intermediate
 recombination} by taking $\mu>1$ individuals into account for a
 weighted average.

The measure\footnote{%
Later, the vector $\ve{w}$ will have $\lam\ge\mu$ elements. Here, for computing the norm, 
we assume that any additional $\lam-\mu$ elements are zero. }
\begin{equation}
  \label{eqmueff}
  \mueff = \kom{\frac{\summ{i}{\mu}w_i}{\summ{i}{\mu}w_i^2} =} 
              \kom{\frac{1}{\summ{i}{\mu}w_i^2}}
            \rklam{\frac{\|\ve{w}\|_1}{\|\ve{w}\|_2}}^2 = 
            \frac{\|\ve{w}\|_1^2}{\|\ve{w}\|_2^2} =  
            \frac{(\summ{i}{\mu}|w_i|)^2}{\summ{i}{\mu}w_i^2} = 
            \frac{1}{\summ{i}{\mu}w_i^2}
\end{equation}
will be repeatedly used in the following and can be paraphrased as \emphindex{effective sample size} of the selected samples or \emphindex{variance effective selection mass}.  From the definition of
$w_i$ in \eqref{eqweights} we derive $1\le\mueff\le\mu$, and
$\mueff=\mu$ for equal recombination weights, \ie\ $w_{i}={1}/{\mu}$
for all $i=1\dots\mu$.  
\begin{einschub}\sloppy
The notion of $\mueff$ with different recombination weights generalizes the notion of $\mu$ with equal recombination weights in several aspects. The number $\mu$ (with equal recombination weights) is the amount of information used, expressed as number of independent sources. 
Taking the weighted average of \emph{independent} samples reduces the original variance by a factor of $\mu$ for equal weights and by a factor of \mueff\ for any weights, hence \mueff\ can be considered as the amount of used information.
To keep the variance unchanged, the average must be multiplied by $\sqrt{\mueff}$.
However, the optimal step-size (given $\mu<\N$) is proportional to $\mu$ \cite{Beyer:2001b,Rechenberg:1994} for equal weights and proportional to $1.25\,\mueff$ for optimal recombination weights \cite{arnold2006weighted}, respectively, see also Section~\ref{secsscontrol}.
\end{einschub}
Usually, $\mueff\approx{\lam}/{4}$ indicates a
reasonable setting of $w_i$. A simple and reasonable setting is
$w_i\propto\mu-i+1$, and $\mu\approx{\lam}/{2}$, where $\mueff\approx3\lam/8$.

The final equation rewrites \eqref{eqreco} as an \emph{update} of \xmean,
\equationframe{{
\begin{equation}\label{eqrecofinal}
  \xmeangg =
  \xmeang + \cm \summ{i}{\mu} w_i\, (\ve{x}_{i:\lam}^{(g+1)} - \xmeang)
\end{equation}
}}
where
\where{$\cm$ is a learning rate, usually set to $1$.\footnote{%
In the literature the notation $\kappa=1/\cm$ is also common and $\kappa$ is used as multiplier in \eqref{eqmut} instead of in \eqref{eqrecofinal}.}}
\whereende
\Eqref{eqrecofinal} generalizes \eqref{eqreco}. If $\cm\summ{i}{\mu}w_i = 1$, as
it is the case with the default parameter setting (compare Table\nref{tabdefpara} in Appendix\nref{sec:algorithm}), $-\xmeang$ cancels out \xmeang, and \Eqsref{eqrecofinal} and \eqrefadd{eqreco} are identical.

\begin{einschub}%
  Choosing $\cm<1$ can be advantageous on noisy functions.
  With optimal step-size we have roughly $\sig\propto1/\cm$, hence the ``test steps'' in \eqref{eqmut} are in effect increased whereas the update step in \eqref{eqrecofinal} remains unchanged.
  However, too large test steps negatively impact the performance because the ranking indices $i\!:\!\lambda$ are determined too far away from the (current) region of relevance.
  Arbitrary small test steps (when $\cm\to\infty$) work generally well, within the limits of numerical precision, on unimodal and noisefree functions \cite{akimoto2017quality,gissler2022learning}.
\end{einschub}
%

\section{Adapting the Covariance Matrix\label{secadaptC}}
In this section, the update of the covariance matrix, $\ma{C}$, is
derived. We will start out estimating the covariance matrix from a
single population of one generation (Sect\pref{sec:estimateC}). For
small populations this estimation is unreliable and an adaptation
procedure has to be invented (rank-$\mu$-update,
Sect\pref{sec:updateC}).  In the limit case only a single point can be
used to update (adapt) the covariance matrix at each generation
(rank-one-update, Sect\pref{sec:rankone}).  This adaptation can be
enhanced by exploiting dependencies between successive steps applying cumulation
(Sect\pref{sec:cumulation}). Finally we combine the rank-$\mu$ and
rank-one updating methods (Sect\pref{sec:combine}).

\subsection{Estimating the Covariance Matrix From Scratch\label{sec:estimateC}}
For the moment we assume that the population contains enough
information to reliably estimate a covariance matrix from the
population.\footnote{%
  To re-estimate the covariance matrix, $\ma{C}$, from a
  $\NormalNullI$ distributed sample such that $\mathtt{cond}(\ma{C})
  <10$ a sample size $\lam\ge4\N$ is needed, as can be observed in 
  numerical experiments. }
For the sake of convenience we assume $\sig^{(g)}= 1$ (see
\eqref{eqmut}) in this section. For $\sigg\not=1$ the formulae hold
except for a constant factor.

 We can (re-)estimate the original covariance matrix $\Cg$ using the
 sampled population from \eqref{eqmut}, $\xgg_1\dots\xgg_\lam$, via
 the empirical covariance matrix
\begin{equation}\label{eqempcomalam}
    \Cgg_\mathrm{emp} = \frac{1}{\lam-1}\sum_{i=1}^{\lam}
    \rklam{\xgg_i - \frac{1}{\lam}\sum_{j=1}^{\lam}\xgg_j}
    \rklam{\xgg_i - \frac{1}{\lam}\sum_{j=1}^{\lam}\xgg_j}^\T
    \enspace.
\end{equation}
The empirical covariance matrix $\Cgg_\mathrm{emp}$ is an
unbiased estimator of $\Cg$:\kom{ \\$\Cgg_\lam = \arg \max_{\ma{C}}
  P(\xgg_1,\dots,\xgg_\lam |\xmeang, \ma{C})$} assuming the $\xgg_{i},
i=1\dots\lam$, to be random variables (rather than a realized sample),
we have that $\Extext{\Cgg_\mathrm{emp}\,\left|\,\Cg\right.} = \Cg$.
Consider now a slightly different approach to get an estimator for
$\Cg$.
\begin{equation}\label{eqestimatelam}
  \Cgg_\lam = \frac{1}{\lam}\sum_{i=1}^\lam 
              \rklam{\xgg_i - \xmeang}\rklam{\xgg_i - \xmeang}^\T
\end{equation}
Also the matrix $\Cgg_\lam$ is an unbiased estimator of $\Cg$. The
remarkable difference between \eqsref{eqempcomalam} and
\eqrefadd{eqestimatelam} is the reference mean value.  For
$\Cgg_\mathrm{emp}$ it is the mean of the \emph{actually realized}
sample.  For $\Cgg_\lam$ it is the \emph{true} mean value, $\xmeang$,
of the sampled distribution (see \eqref{eqmut}).  Therefore, the
estimators $\Cgg_\mathrm{emp}$ and $\Cgg_\lam$ can be interpreted
differently: while $\Cgg_\mathrm{emp}$ estimates the distribution
variance \emph{within the sampled points}, $\Cgg_\lam$ estimates
variances of sampled \emph{steps}, $\xgg_i - \xmeang$. 
\begin{einschub}%
 A minor difference between \eqsref{eqempcomalam} and
 \eqrefadd{eqestimatelam} is the different normalizations
 $\frac{1}{\lam-1}$ versus $\frac{1}{\lam}$, necessary to get an
 unbiased estimator in both cases. In \eqsref{eqempcomalam} one degree
 of freedom is already taken by the inner summand. In order to get a
 \emph{maximum likelihood} estimator\kom{ for the $\xgg_i-\xmeang$}, 
 in both cases $\frac{1}{\lam}$ must be
 used.
\end{einschub}

\Eqref{eqestimatelam} re-estimates \emph{the original} covariance
matrix.  To ``estimate'' a ``better'' covariance matrix,
the same, \emph{weighted
  selection} mechanism as in \eqref{eqreco} is used
\cite{Hansen:2004b}.
\begin{equation}\label{eqCmu}
  \Cgg_\mu = \summ{i}{\mu} w_i \rklam{\xgg_{i:\lam} - \xmeang}
      \rklam{\xgg_{i:\lam} - \xmeang}^\T
\end{equation}
The matrix $\Cgg_\mu$is an estimator for the distribution of
\emph{selected steps}, just as $\Cgg_\lam$ is an estimator of the
original distribution of steps before selection. Sampling from
$\Cgg_\mu$ tends to reproduce selected, \ie\ \emph{successful} steps,
giving a justification for what a ``better'' covariance
matrix means. 

{%

\begin{einschub}%
  Following \cite{Hansen:2006}, we compare \eqref{eqCmu} with the
  Estimation of Multivariate Normal Algorithm
  EMNA$_{global}$\index{EMNA}
  \cite{Larranaga:2002,Larranaga:2001}. The covariance matrix in
  EMNA$_{global}$ reads, similar to \eqsref{eqempcomalam},
  \begin{equation}\label{eqCemna}
  \Cgg_{\mathrm{EMNA}_{global}} = 
      \frac{1}{\mu}\summ{i}{\mu} \rklam{\xgg_{i:\lam} - \xmeangg}
      \rklam{\xgg_{i:\lam} - \xmeangg}^\T \enspace,
  \end{equation}
  where $\xmeangg=\frac{1}{\mu}\summ{i}{\mu}\xgg_{i:\lam}$.{
  Similarly, applying the so-called Cross-Entropy method\index{Cross-Entropy method} to continuous
  domain optimization \cite{Rubenstein:2004} yields the covariance
  matrix $\frac{\mu}{\mu-1}\, \Cgg_{\mathrm{EMNA}_{global}}$, \ie\ the
  \emph{unbiased} empirical covariance matrix of the $\mu$ best points.}
  In both cases the subtle\tutorial{, but most important} difference
  to \eqref{eqCmu} is, again, the choice of the reference mean
  value.\footnote{%
    Taking a weighted sum, $\summ{i}{\mu}w_i\dots$, instead of the mean,
     $\frac{1}{\mu}\summ{i}{\mu}\dots$, is an appealing, but less important,
     difference.}
   \Eqref{eqCemna} estimates the {variance \emph{within} the selected
   population} while \eqref{eqCmu} estimates {selected steps}.
   \Eqref{eqCemna} reveals always smaller variances than
   \eqref{eqCmu}, because its reference mean value is the minimizer
   for the variances.  Moreover, in most conceivable selection
   situations \eqref{eqCemna} decreases the variances compared to
   $\Cg$.

\begin{figure}[t]
  \begin{center}
    \begin{minipage}{0.99\textwidth} 
{\parbox[t]{0.99\textwidth}{
      \includegraphics[width=0.3574\textwidth]%
                                      {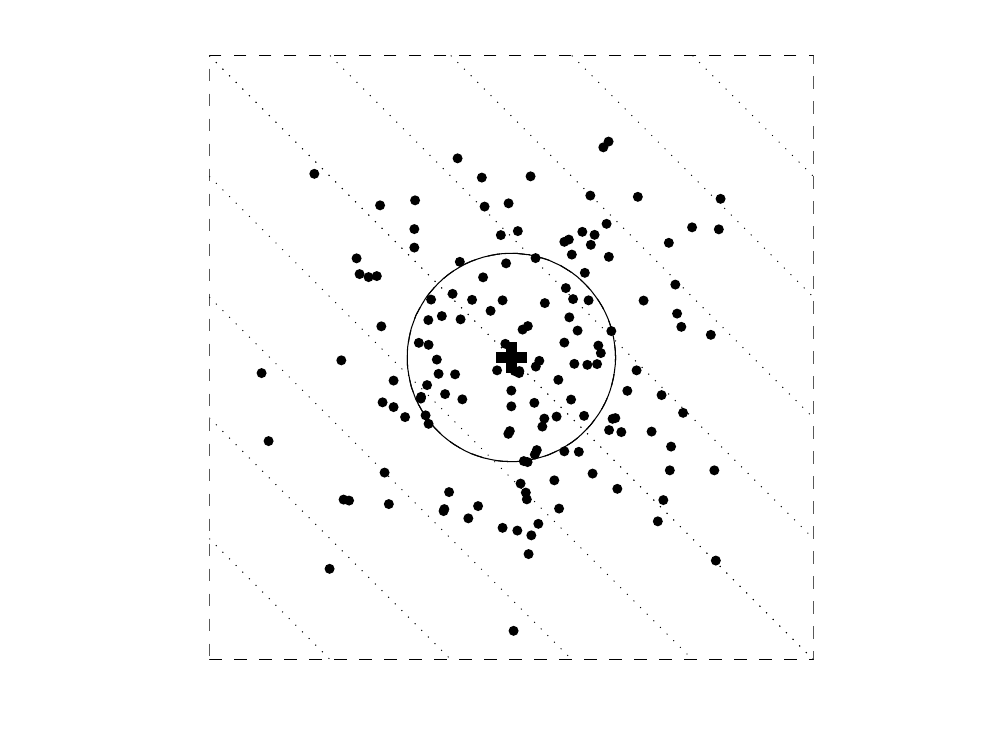}%
      \hspace*{-0.067\textwidth}
      \includegraphics[width=0.3574\textwidth]%
                                      {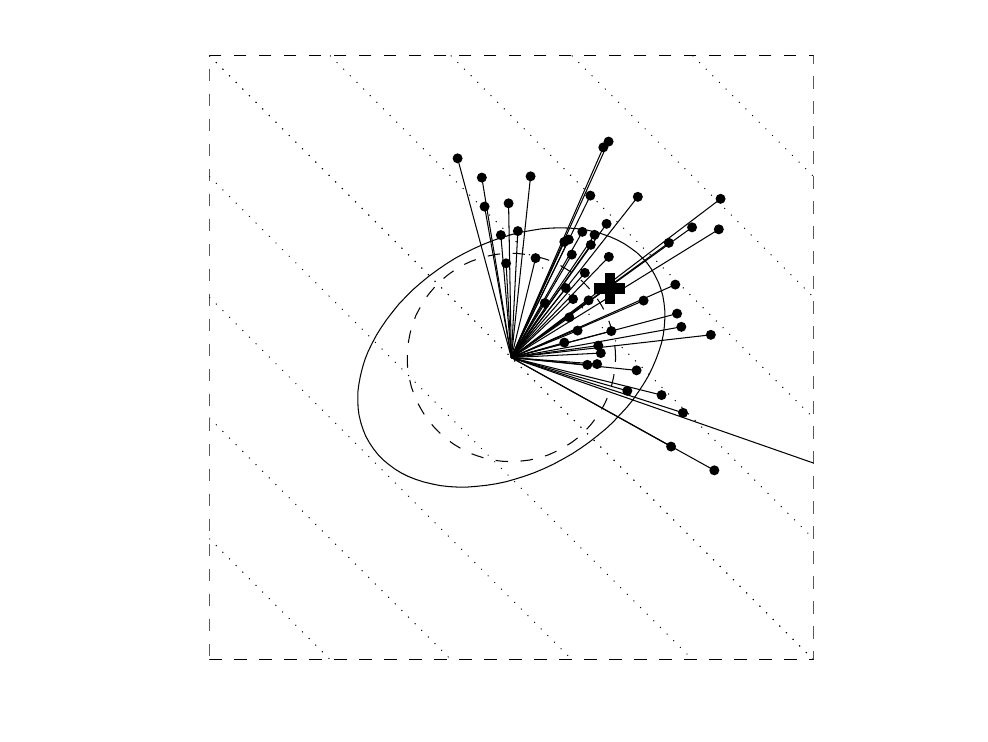}%
      \hspace*{-0.067\textwidth}
      \includegraphics[width=0.3574\textwidth]%
                                      {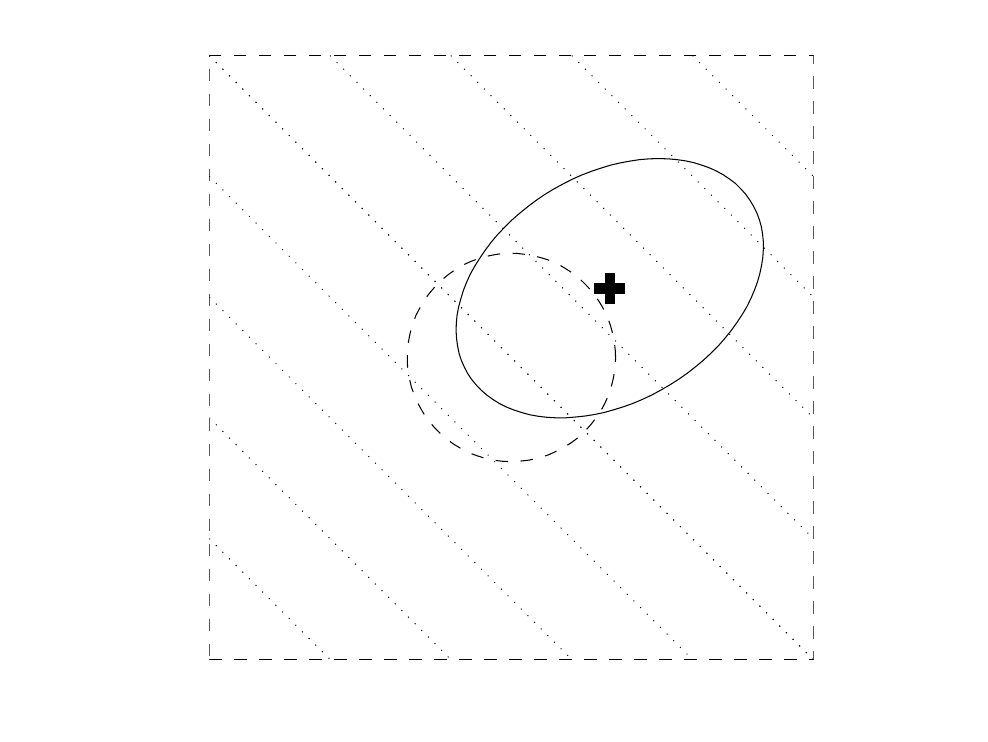}%
      \raisebox{10ex}{$\Cgg_\mu$}
\\[-3ex]
}}
{\parbox[t]{0.99\textwidth}{ 
      \includegraphics[width=0.3574\textwidth]%
                                      {Figures/sampleorig}%
      \hspace*{-0.067\textwidth}
      \includegraphics[width=0.3574\textwidth]%
                                      {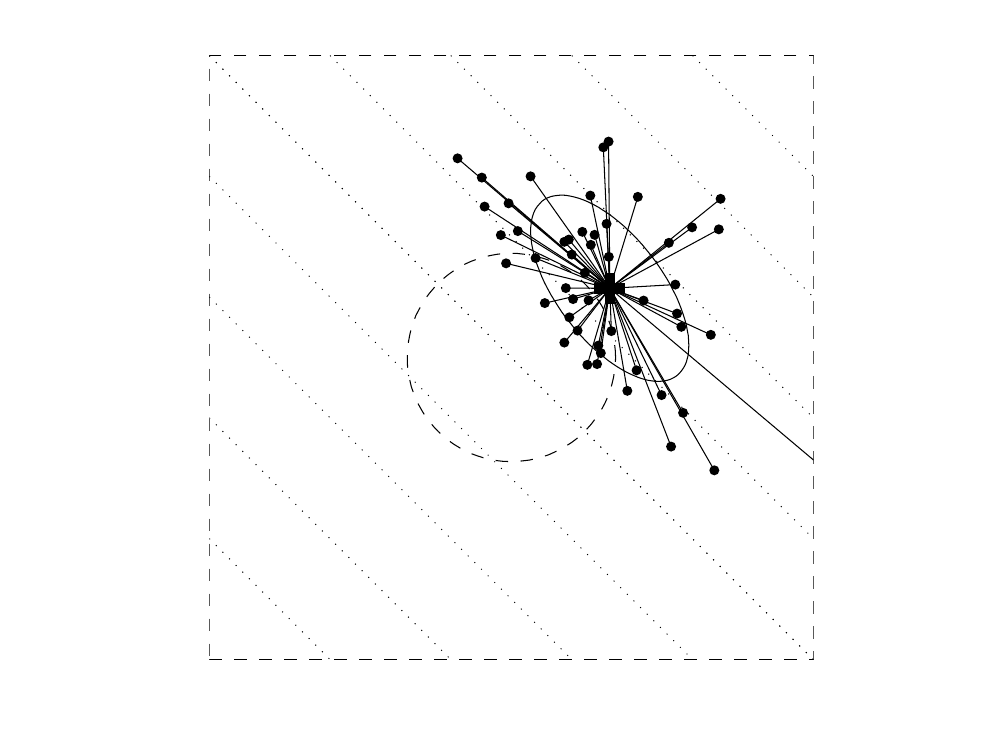}%
      \hspace*{-0.067\textwidth}
      \includegraphics[width=0.3574\textwidth]%
                                      {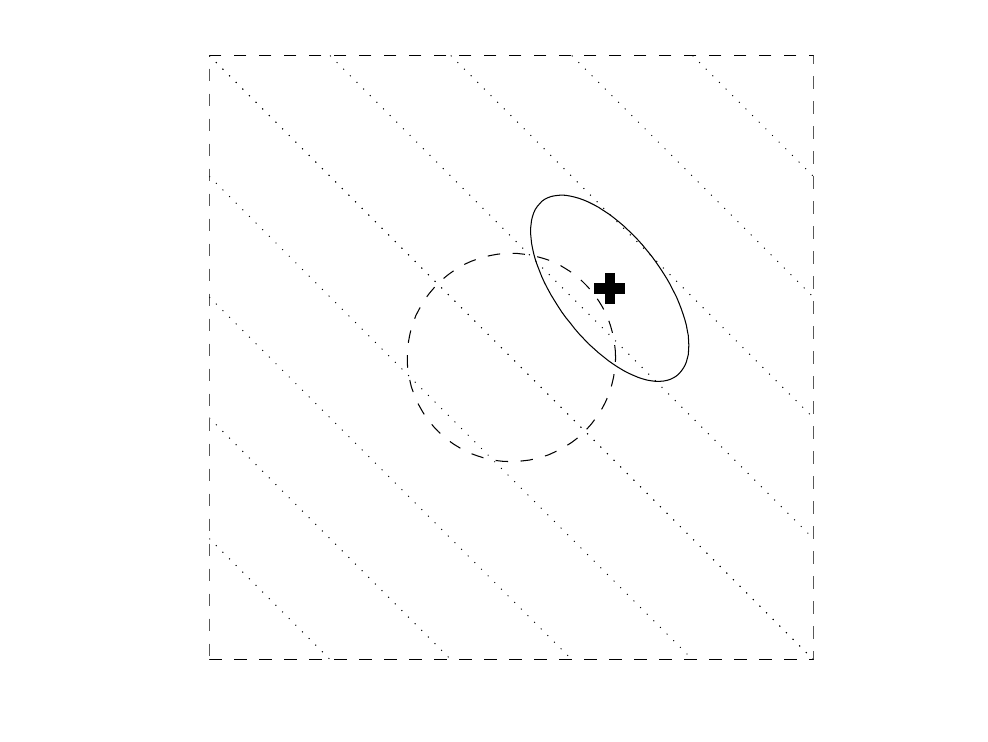}%
      \raisebox{10ex}{$\Cgg_{\mathrm{EMNA}_{global}}$}
}}
\parbox{0.31\textwidth}{\centering 
sampling 
}
\parbox{0.31\textwidth}{\centering 
estimation
}
\parbox{0.31\textwidth}{\centering 
new distribution
}
    \end{minipage}
  \end{center}
\caption[simpleeda]{\label{fig:simpleeda}Estimation of the covariance
  matrix on $\flinear(\ve{x})=-\summ{i}{2} x_i$ to be minimized.
  Contour lines (\textit{dotted}) indicate that the strategy should
  move toward the upper right corner.  \auge{Above}: estimation of
  $\Cgg_\mu$ according to \eqref{eqCmu}, where $w_i=1/\mu$.
  \auge{Below}: estimation{ of $\Cgg_{\mathrm{EMNA}_{global}}$}
  according to \eqref{eqCemna}. Left: sample of $\lam=150$
  $\NormalNullI$ distributed points.  Middle: the $\mu=50$ selected
  points (\textit{dots}) determining the entries for the estimation
  equation (\textit{solid straight lines}). Right: search distribution
  of the next generation (\textit{solid ellipsoids}).  Given $w_i=1/\mu$, estimation via
  $\Cgg_\mu$ \emph{increases} the expected variance in gradient
  direction for all $\mu<\lam/2$, while estimation via
  $\Cgg_{\mathrm{EMNA}_{global}}$ \emph{decreases} this variance for
  any $\mu<\lam$ geometrically fast}
\end{figure}
 Figure\nref{fig:simpleeda} demonstrates the estimation results on
 \emph{a linear} objective function for $\lam=150$, $\mu=50$, and
 $w_i=1/\mu$.  \Eqref{eqCmu} geometrically increases the expected
 variance in direction of the gradient (where the selection takes
 place, here the diagonal), given ordinary settings for parent number
 $\mu$ and recombination weights $w_1,\dots,w_\mu$. \Eqref{eqCemna}
 always decreases the variance in gradient direction geometrically
 fast!  Therefore, \eqref{eqCemna} is highly susceptible\kom{prone} to
 premature convergence, in particular with small parent populations,
 where the population cannot be expected to bracket the optimum at any
 time.  However, for large values of $\mu$ in large populations with
 large initial variances, the impact of the different reference mean
 value can become marginal.
\end{einschub}}

%
 In order to ensure with \eqsref{eqmut}, \eqrefadd{eqreco}, and
 \eqrefadd{eqCmu}, that $\Cgg_\mu$ is a \emph{reliable} estimator, the
 variance effective selection mass $\mueff$ (cf.\ \eqref{eqmueff})
 must be large enough: getting condition numbers (cf.\
 Sect\pref{def:condition}) smaller than ten for $\Cg_\mu$ on
 $\fsphere(\ve{x})=\summ{i}{\N}x_i^2$, requires $\mueff\approx10\N$.
\kom{AR smaller 3 for re-estimation $3\N$ is sufficient}The next step
is to circumvent this restriction on $\mueff$.

\subsection{Rank-$\mu$-Update\label{sec:updateC}}
To achieve \emph{fast} search (opposite to  \emph{more robust} or
\emph{more global} search), e.g.\ competitive performance on $\fsphere:\x\mapsto\sum x_i^2$,
the population size $\lam$ must be small\kom{ (e.g.\
$\lam\le5+2\,\ln\N$)}. Because typically (and ideally) $\mueff\approx\lam/4$ also $\mueff$
must be small and we may assume, e.g., $\mueff\le 1+\ln\N$. Then, it
is not possible to get a \emph{reliable} estimator for a good
covariance matrix from \eqref{eqCmu}.  As a remedy, information from
previous generations is used additionally.  For example, after a
sufficient number of generations, the mean of the estimated covariance
matrices from all generations,
\begin{equation}
  \Cgg =
  \frac{1}{g+1}\sum_{i=0}^{g}\frac{1}{{\sig^{(i)}}^2}\ma{C}^{(i+1)}_\mu
  \label{eqallCs} 
\end{equation}
becomes a reliable estimator for the selected steps. To make $\Cg_\mu$
from different generations comparable, the different $\sig^{(i)}$ are
incorporated.{ (Assuming $\sig^{(i)}=1$, \eqref{eqallCs} resembles the
  covariance matrix from the Estimation of Multivariate Normal Algorithm
  EMNA$_i$ \cite{Larranaga:2001}.)}

In \eqref{eqallCs}, all generation steps have the same weight. To
assign recent generations a higher weight, exponential smoothing is
introduced. Choosing $\ma{C}^{(0)}=\Id$ to be the unity matrix and a
learning rate $0<\cmu\le1$, then $\Cgg$ reads
\begin{eqnarray}
  \Cgg &=& (1-\cmu)\Cg + \cmu\frac{1}{{\sigg}^2}\Cgg_\mu \nonumber\\
       &=& (1-\cmu)\Cg + \cmu\summ{i}{\mu}w_i
            \,\ygg_\ilam {\ygg_\ilam}^\T \label{eqrankmupositive}
            \enspace,
\end{eqnarray}
where 
\where{$\cmu\le 1$ learning rate for updating the covariance
  matrix. For $\cmu=1$, no prior information is retained and
  $\Cgg=\frac{1}{{\sigg}^2}\Cgg_\mu$. For $\cmu=0$, no learning takes
  place and $\Cgg=\ma{C}^{(0)}$. Here, $\cmu\approx\min(1,\mueff/\N^2)$
  is a reasonably choice.
}
\where{$w_{1\dots\mu}\in\R$ such that $w_1\ge\dots\ge w_\mu> 0$ and $\sum_i w_i = 1$.} 
\where{$\ygg_\ilam = {(\xgg_{i:\lam} - \xmeang)}/{\sigg}$. }
\where{$\zgg_\ilam = {\Cg}^{-1/2} \ygg_\ilam$ is the mutation vector expressed in the unique coordinate system where the sampling is isotropic and the respective coordinate system transformation does not rotate the original principal axes of the distribution.}
\whereende
This covariance matrix update is called rank-$\mu$-update
\cite{Hansen:2003}, because the sum of outer products in
\eqref{eqrankmupositive} is of rank $\min(\mu,\N)$ with probability one (given 
$\mu$ non-zero weights).
This sum can even consist of a single term, if $\mu=1$.

Finally, we generalize \eqref{eqrankmupositive} to $\lambda$ weight values which need neither sum to $1$, nor be non-negative anymore \cite{jastrebski2006improving, hansen2010benchmarking}, 
\begin{eqnarray}\label{}
    \Cgg &=& \left(1-\cmu\sumw[i]\right)\Cg + \cmu\summ{i}{\lam}w_i
            \ygg_\ilam {\ygg_\ilam}^\T \label{eqrankmu} \\
       &=& {\Cg}^{1/2} \left( \Id + \cmu \summ{i}{\lam}w_i
            \left(\zgg_\ilam {\zgg_\ilam}^\T - \Id\right)\right) {\Cg}^{1/2}
  \enspace,\nonumber  
\end{eqnarray}
where
\where{$w_{1\dots\lam}\in\R$ such that $w_1\ge\dots\ge w_\mu> 0 \ge w_{\mu+1} \ge w_\lam$, and usually $\summ{i}{\mu} w_i = 1$ and $\summ{i}{\lam} w_i \approx 0$.
}
\where{$\sumw[i]=\summ{i}{\lam}w_i$}
\whereende
The second line of \eqref{eqrankmu} expresses the update in the natural coordinate system, an idea already considered in \cite{glasmachers2010exponential}. The identity covariance matrix is updated and a coordinate system transformation is applied afterwards by multiplication with ${\Cg}^{1/2}$ on both sides.
\Eqref{eqrankmu} uses $\lam$ weights, $w_i$, of which about half are negative. If the weights are chosen such that $\sumw[i] = 0$, the decay on $\Cg$ disappears and changes are only made along axes in which samples are realized. 
\begin{einschub}%
Negative values for the recombination weights in the covariance matrix update have been introduced in the seminal paper of Jastrebski and Arnold \cite{jastrebski2006improving} as \emph{active} covariance matrix adaptation. Non-equal negative weight values have been used in \cite{hansen2010benchmarking} together with a rather involved mechanism to make up for different vector lengths. The default recombination weights as defined in Table\nref{tabdefpara} in Appendix\nref{sec:algorithm} are somewhere in between these two proposals, but closer to \cite{jastrebski2006improving}. Slightly deviating from \eqref{eqrankmu} later on, vector lengths associated with negative weights will be rescaled to a (direction dependent) constant, see \eqref{eq-def-w-dyn} and \eqrefadd{algcov} in Appendix\nref{sec:algorithm}. This allows to \emph{guaranty} positive definiteness of $\Cgg$. Conveniently, it also alleviates a selection error which usually makes directions associated with longer vectors worse. 
\end{einschub}
The number ${1}/{\cmu}$ is the \auge{backward time horizon}
that contributes roughly $63\%$ of the overall information.\kom{ that is, $\Cgg$
  is a weighted mean of{ the $g+2$ matrices} $\ma{C}^{(0)}$,
  $\frac{1}{{\sig^{(0)}}^2}\ma{C}^{(1)}_\mu$,
  $\frac{1}{{\sig^{(1)}}^2}\ma{C}^{(2)}_\mu,\dots,$
  $\frac{1}{{\sig^{(g)}}^2}\Cgg_\mu$.}%
\begin{einschub}%
  Because \eqref{eqrankmu} expands to the weighted sum
  \begin{equation}
  \Cgg = (1-\cmu)^{g+1}\ma{C}^{(0)} + 
     \cmu\suml{i}{g}{(1-\cmu)^{g-i}}
     \frac{1}{{\sig^{(i)}}^2}\;\Cii_\mu \enspace,
  \end{equation}
  the backward time horizon, $\rmDelta g$, where about $63\%$ of the
  overall weight is summed up, is defined by
  \begin{equation}
  \cmu\sum_{i=g+1-\rmDelta g}^{g}{(1-\cmu)^{g-i}} \approx 0.63 \approx 1 -
  \frac{1}{\mathrm{e}} \enspace.
  \end{equation}
Resolving the sum yields
{\begin{equation}\label{eqweightreduction}
  (1-\cmu)^{\rmDelta g} \approx \frac{1}{\mathrm{e}} \enspace,
\end{equation}
and resolving for $\rmDelta g$, using the Taylor approximation for
$\ln$, yields
\begin{equation}\label{eqcovtime}
   \rmDelta g\approx\frac{1}{\cmu} \enspace.
\end{equation}
}%
That is, approximately $37\%$ of the information in $\Cgg$ is older
than ${1}/{\cmu}$ generations, and, according to
\eqref{eqweightreduction}, the original weight is reduced by a factor
of $0.37\kom{\approx\frac{1}{2.7} \approx\frac{1}{e}}$ after
approximately ${1}/{\cmu}$ generations.\footnote{%
   This can be shown more easily, because $(1-\cmu)^g =
   \exp\ln(1-\cmu)^g = \exp(g\ln(1-\cmu)) \approx \exp(-g\cmu)$ for
   small $\cmu$, and for $g\approx 1/\cmu$ we get immediately
   $(1-\cmu)^g\approx\exp(-1)$.}

\end{einschub}
The choice of $\cmu$ is crucial. Small values lead to slow
learning, too large values lead to a failure, because the covariance
matrix degenerates. Fortunately, a good setting seems to be largely
independent of the function to be optimized.\footnote{%
  We use the sphere model $\fsphere(\ve{x})=\sum_i x_i^2$ to
  empirically find a good setting for the parameter $\cmu$, dependent
  on $\N$ and $\mueff$.  The found setting was applicable to any
  non-noisy objective function we tried so far. }
A first order approximation for a good choice is
$\cmu\approx{\mueff}/{\N^2}$.  Therefore, the characteristic time
horizon for \eqref{eqrankmu} is roughly ${\N^2}/{\mueff}$. 

Experiments 
suggest that $\cmu\approx\mueff/\N^2$ is a rather conservative setting for large values of \N, 
whereas $\mueff/\N^{1.5}$ appears to be slightly beyond the limit of  
stability. The best, yet robust choice of the exponent remains to 
be an open question. 

\label{text:learninganditerations}
\newcommand{\reflearninganditerations}{end of
  Section\nref{text:learninganditerations}} Even for the learning rate
$\cmu=1$, adapting the covariance matrix cannot be accomplished
within one generation. The effect of the original sample distribution
does not vanish until a sufficient number of generations.  Assuming fixed
search costs (number of function evaluations), a small population size
$\lam$ allows a larger number of generations and therefore usually
leads to a faster adaptation of the covariance matrix.

\newcommand{\new}[1]{}
\renewcommand{\new}[1]{#1}

\subsection{Rank-One-Update\label{sec:rankone}}

 In Section\nref{sec:estimateC} we started by estimating the complete covariance
 matrix from scratch, using all selected steps from a \emph{single
 generation}. We now take an opposite viewpoint. We
 repeatedly update the covariance matrix in the generation
 sequence using a \emph{single selected step} only. First, this
 perspective will give another justification of the adaptation rule
 \eqref{eqrankmu}. Second, we will introduce the so-called evolution
 path that is finally used for a rank-one update of the covariance matrix.

\subsubsection{A Different Viewpoint}
We consider a specific method to produce $\N$-dimensional normal
 distributions with zero mean. Let the vectors
 $\ve{y}_1,\dots,\ve{y}_{g_0}\in\Rn$, $g_0\ge \N$, span $\Rn$ and let
 $\Normal{0,1}$ denote independent $(0,1)$-normally distributed random
 numbers, then
\begin{equation}\label{eqsumlinien}
        \Normal{0,1}\ve{y}_1 + \dots + \Normal{0,1}\ve{y}_{g_0}
       \;\;\sim\;\;\Normal{\ve{0},\sum_{i=1}^{g_0} \ve{y}_i\ve{y}_i^\T}
\end{equation}
 is a normally distributed random vector with zero
 mean and covariance matrix
   $\kom{\ma{C} = }\sum_{i=1}^{g_0} \ve{y}_i\ve{y}_i^\T$.
%
%
 The random vector \eqref{eqsumlinien} is generated by adding
 ``line-distributions''
 $\Normal{0,1}\ve{y}_i\kom{\,\sim\textNormal{\ve{0},\ve{y}_i\ve{y}_i^{\T}}}$.
 The singular distribution $\Normal{0,1}\ve{y}_i \sim
 \textNormal{\ve{0},\ve{y}_i\ve{y}_i^{\T}}$ generates the vector
 $\ve{y}_{i}$ with maximum likelihood considering all normal
 {distributions} with zero mean.
 \begin{einschub}%
   The line distribution that generates a vector $\ve{y}$ with the
   maximum likelihood must ``live'' on a line that includes $\ve{y}$,
   and therefore the distribution must obey $\textNormal{0, 1} \sigma
   \ve{y} \sim \textNormal{0, \sigma^2\ve{y}\ve{y}^\T}$. Any other
   line distribution with zero mean cannot generate $\ve{y}$ at
   all. Choosing $\sigma$ reduces to choosing the maximum likelihood
   of $\|\ve{y}\|$ for the one-dimensional gaussian
   $\textNormal{0,\sigma^2\|\ve{y}\|^2}$, which is $\sigma=1$.

   The covariance matrix $\ve{y}\ve{y}^\T$ has rank one, its only
   eigenvectors are $\{\alpha\ve{y}\,|\, \alpha\in\R_{\setminus 0}\}$ with eigenvalue
   $\|\ve{y}\|^2$.  Using \eqeqref{eqsumlinien}, any normal
   distribution can be realized if $\ve{y}_i$ are chosen
   appropriately. For example, \eqref{eqsumlinien} resembles
   \eqref{eq:normalforms} with $\ve{m}=\ve{0}$, using the orthogonal
   eigenvectors $\ve{y}_i=d_{ii}\ve{b}_i$, for $i=1,\dots,\N$, where
   $\ve{b}_i$ are the columns of $\ma{B}$. In general, the vectors
   $\ve{y}_i$ need not to be eigenvectors of the covariance matrix,
   and they usually are not. \kom{ Moreover, given \eqeqref{eqCsumy} for
   the covariance matrix we can easily generate the distribution using
   \eqref{eqsumlinien}.}
\end{einschub}

 Considering \eqref{eqsumlinien} and a slight simplification of
 \eqref{eqrankmu}, we try to gain insight into the adaptation rule for
 the covariance matrix. Let the sum in \eqref{eqrankmu} consist of a
 single summand only (\eg\ $\mu=1$), and let $\ve{y}_{g+1} =
 \frac{\xgg_{1:\lam} - \xmeang}{\sigg}$.  Then, the rank-one update for the
 covariance matrix reads
\begin{eqnarray}
  \Cgg &=& (1-\cone)\Cg + \cone \,{\ve{y}_{g+1}} {\ve{y}_{g+1}}^\T
       \label{eqrankonesimp}
\end{eqnarray}
 The right summand is of rank one and adds the maximum likelihood term
 for $\ve{y}_{g+1}$ into the covariance matrix $\Cg$. Therefore the
 probability to generate $\ve{y}_{g+1}$ in the next generation
 increases.\kom{%
   We believe that $\Cgg$ achieves maximum likelihood for
   $\ve{y}_{g+1}$, given the prior $\Cg$ and zero mean, but the exact
   conditions have yet to been shown. }

\begin{figure}[tb]
  \begin{center}
      \includegraphics[width=0.32409\textwidth]{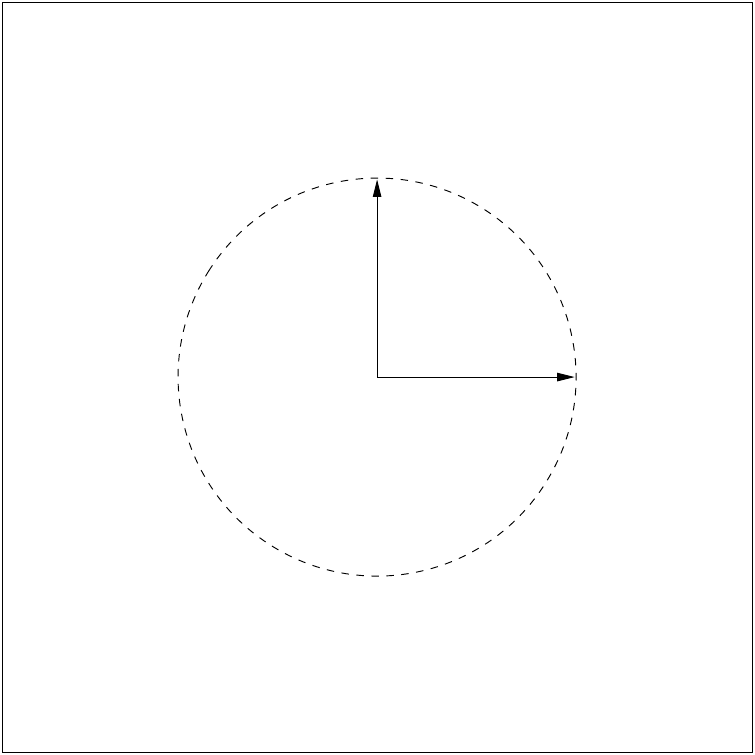}
      \includegraphics[width=0.32409\textwidth]{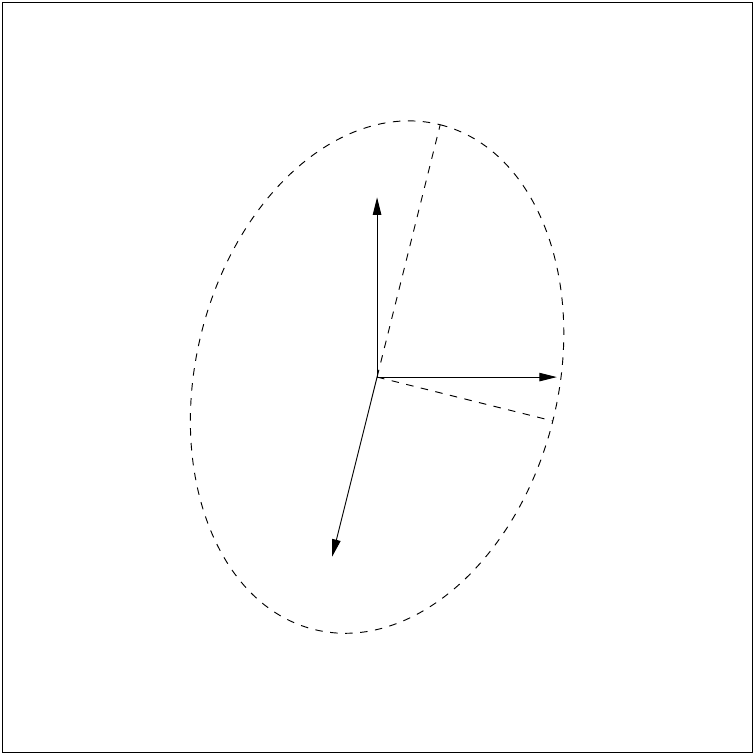}
      \includegraphics[width=0.32409\textwidth]{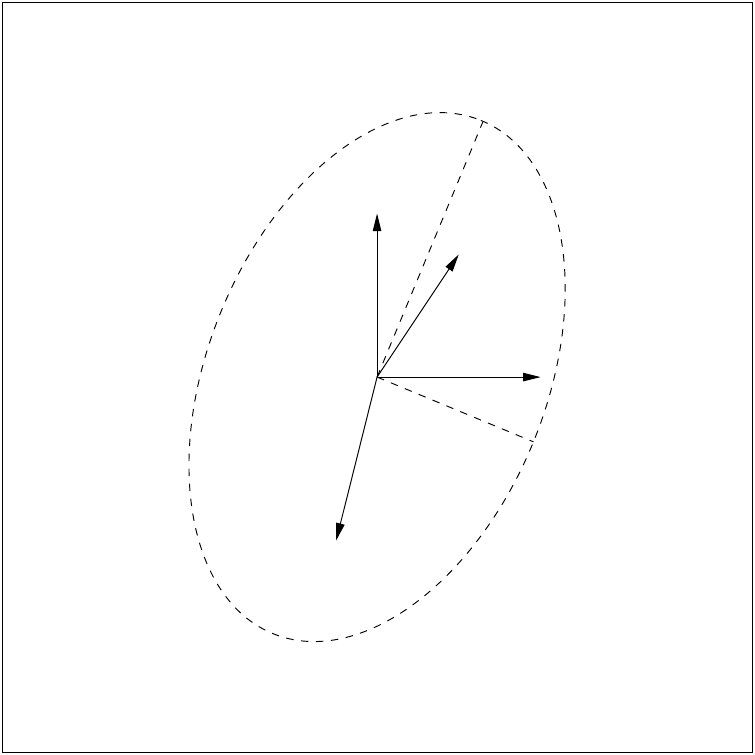}\\
\parbox{0.33\textwidth}{ 
\centerline{$\Normal{\ve{0},\ma{C}^{(0)}}$}%
}
\parbox{0.33\textwidth}{ 
\centerline{$\Normal{\ve{0},\ma{C}^{(1)}}$}%
}
\parbox{0.30\textwidth}{ 
\centerline{$\Normal{\ve{0},\ma{C}^{(2)}}$}
}
\vspace{-2ex}
  \end{center}
  \caption[construct]{\label{figvertkonst}Change of the distribution
  according to the covariance matrix update \eqref{eqrankonesimp}\kom{
  using the distribution construction principle from
  \eqref{eqsumlinien}}. Left: vectors $\ve{e}_1$ and $\ve{e}_2$, and
  $\ma{C}^{(0)}=\Id=\ve{e}_1\ve{e}_1^\T + \ve{e}_2\ve{e}_2^\T$.
  Middle: vectors $0.91\,\ve{e}_1$, $0.91\,\ve{e}_2$, and
  $0.41\,\ve{y}_1$ (the coefficients deduce from $\cone=0.17$), and
  $\ma{C}^{(1)}=(1-\cone)\,\Id + \cone\,\ve{y}_1\ve{y}_1^\T $, where
  $\ve{y}_1 = {-0.59\choose -2.2}$. The distribution ellipsoid is
  elongated into the direction of $\ve{y}_1$, and therefore increases
  the likelihood of $\ve{y}_1$. Right:
  $\ma{C}^{(2)}=(1-\cone)\,\ma{C}^{(1)} + \cone\,\ve{y}_2\ve{y}_2^\T$,
  where $\ve{y}_2 = \rklam{0.97\atop 1.5}$.
  \kom{$\cone=2/(2+sqrt(2))^2=0.1716$} }
\end{figure}

 An example of the first two iteration steps of \eqref{eqrankonesimp}
 is shown in \auge{Figure\nref{figvertkonst}}.%
 The distribution $\textNormal{\ve{0},\ma{C}^{(1)}}$ tends to
 reproduce $\ve{y}_{1}$ with a larger probability than the initial
 distribution $\textNormal{\ve{0},\Id}$; the distribution
 $\textNormal{\ve{0},\ma{C}^{(2)}}$ tends to reproduce $\ve{y}_{2}$
 with a larger probability than $\textNormal{\ve{0},\ma{C}^{(1)}}$,
 and so forth. When $\ve{y}_1,\dots,\ve{y}_g$ denote the formerly
 selected, favorable steps, $\textNormal{\ve{0},\Cg}$ tends to
 reproduce these steps. The process leads to an alignment of the
 search distribution $\textNormal{\ve{0},\Cg}$ to the distribution of
 the selected steps. If both distributions become alike, as under
 random selection, in expectation no further change of the covariance
 matrix takes place \cite{Hansen:1998}.

\subsubsection{Cumulation: Utilizing the Evolution Path\label{sec:cumulation}}
 We have used the selected steps, $\ygg_\ilam=({\xgg_\ilam} -
 \xmeang)/{\sigg}$, to update the covariance matrix in
 \eqsref{eqrankmu} and \eqrefadd{eqrankonesimp}. Because $\y\y^\T =
 -\y(-\y)^\T$, \emph{the sign of the steps is irrelevant} for the
 update of the covariance matrix\gstrich{}that is, the sign
 information is lost when calculating $\Cgg$. To reintroduce the sign
 information, a so-called \emphindex{evolution path} 
 is constructed \cite{Hansen:1996,Hansen:2001}.

We call a sequence of successive steps, the strategy takes over a
number of generations, an {evolution path}.  An evolution path can be
expressed by a sum of consecutive steps.  This summation is referred
to as \emphindex{cumulation}. To construct an evolution path, the
step-size $\sigma$ is disregarded. For example, an evolution path of
three steps of the distribution mean $\xmean$ can be constructed by
the sum
\begin{equation}\label{eqcumrough}
  \frac{\xmeangg-\xmeang}{\sigg} +
  \frac{\xmeang-\xmean^{(g-1)}}{\sig^{(g-1)}} +
  \frac{\xmean^{(g-1)}-\xmean^{(g-2)}}{\sig^{(g-2)}} \enspace. 
\end{equation}
%
%
 In practice, to construct the evolution path, $\pc\in\Rn$, we use
 exponential smoothing as in \eqref{eqrankmu}, and start with
 $\pc^{(0)}=\ve{0}$.\footnote{%
    In the final algorithm \eqref{eqpc} is still slightly modified,
    compare \eqref{algpc}. }
\equationframe{{
\begin{equation}\label{eqpc}
  \pcgg = (1-\cc)\pcg + \sqrt{\cc(2-\cc)\mueff}\; 
      \frac{\xmeangg-\xmeang}{\cm\sigg} 
\end{equation}
}}
where
\where{$\pcg\in\Rn$, evolution path at generation $g$.}
\where{$\cc\le1$. Again, ${1}/{\cc}$ is the backward time horizon of
  the evolution path $\pc$ that contains roughly $63\%$ of the overall
  weight (compare derivation of \eqref{eqcovtime}). A time horizon
  between $\sqrt{\N}$ and $\N$ is effective. }
\whereende
The factor $\sqrt{\cc(2-\cc)\mueff}$ is a normalization constant for
$\pc$. For $\cc = 1$ and $\mueff=1$, the factor reduces to one, and
$\pcgg={(\xgg_{1:\lam}-\xmeang)}/{\sigg}$.
\begin{einschub}%
  The factor $\sqrt{\cc(2-\cc)\mueff}$ is chosen, such that
\begin{equation}\label{eqpcdist}
  \pcgg\sim\Normal{\ve{0}, \ma{C}}
\end{equation}
if
\begin{equation}\label{eqpcdistassum}
   \pcg\sim\frac{\xgg_{i:\lam}-\xmeang}{\sigg}\sim\Normal{\ve{0},\ma{C}}
   \quad\mbox{for all $i=1,\dots,\mu$ \enspace.} 
\end{equation}
To derive \eqref{eqpcdist} from \eqsref{eqpcdistassum} and
\eqrefadd{eqpc} remark that
  \begin{equation}
    (1-\cc)^2 + \sqrt{\cc(2-\cc)}^2 = 1\quad\mbox{and}\quad
    \summ{i}{\mu}w_i\Normali[i]{\ve{0},\ma{C}}\sim\frac{1}{\sqrt{\mueff}}\Normali{\ve{0},\ma{C}} \enspace.
  \end{equation}
\end{einschub}
The (rank-one) update of the covariance matrix $\Cg$ via the
evolution path $\pcgg$ reads \cite{Hansen:1996}
\begin{equation}\label{eqrankone}
  \Cgg = (1-\cone)\Cg + \cone\pcgg{\pcgg}^\T \enspace.
\end{equation}
An empirically validated choice for the learning rate in
\eqref{eqrankone} is $\cone\approx{2}/{\N^2}$. For $\cc=1$ and
$\mu=1$, \Eqsref{eqrankone}, \eqref{eqrankonesimp}, and
\eqref{eqrankmu} are identical.

Using the evolution path for the update of $\ma{C}$ is a significant
improvement of \eqref{eqrankmu} for small $\mueff$, because
correlations between consecutive steps are heavily exploited. The leading
signs of steps, and the dependencies between consecutive steps play a
significant role for the resulting evolution path $\pcgg$.
\begin{einschub}%
We consider the two most extreme situations, fully correlated steps and entirely anti-correlated steps. The summation in \eqref{eqpc} reads for positive correlations
$$
\sum_{i=0}^\gen (1-\cc)^i \to \frac{1}{\cc} \quad\text{(for $\gen\to\infty$)}\enspace,
$$
and for negative correlations
\begin{eqnarray*}
\sum_{i=0}^\gen (-1)^i (1-\cc)^i 
&=& \sum_{i=0}^{\lfloor\gen/2\rfloor} (1-\cc)^{2i} - \sum_{i=0}^{(\gen-1)/2} (1-\cc)^{2i+1} 
\\
&=& \sum_{i=0}^{\lfloor\gen/2\rfloor} (1-\cc)^{2i} - (1-\cc)\!\!\sum_{i=0}^{(\gen-1)/2} (1-\cc)^{2i}
\\ 
&=& \cc \sum_{i=0}^{\lfloor\gen/2\rfloor} \rklam{(1-\cc)^2}^i 
  + (1-\cc)^\gen ((g + 1) \bmod 2) 
\\
&\to& \frac{\cc}{1-(1-\cc)^2} = \frac{1}{2-\cc} \quad\text{(for $\gen\to\infty$)}\enspace.
\end{eqnarray*}%
\newcommand{\fracc}[2]{#1/#2}%
Multipling these by $\sqrt{\cc(2-\cc)}$, which is 
applied to each input vector, we find that the length of the evolution
path is modulated by the factor of up to 
\begin{align}\label{eq-path-effect}
\sqrt{\frac{2-\cc}{\cc}} &\approx \frac{1}{\sqrt{\cc}}
\end{align}
due to the positive correlations, or its inverse due to negative correlations, respectively \cite[Equations (48) and (49)]{hansen2014principled}.  
\end{einschub}%
With $\sqrt{\N}\le1/\cc\le\N/2$ the number of function evaluations needed to adapt a
nearly optimal covariance matrix on cigar-like objective functions
becomes ${\cal O}(\N)$, despite a learning rate of $\cone\approx2/\N^2$ \cite{hansen2014principled}. A plausible interpretation of this effect is two-fold. First, the desired axis is represented in the path (much) more accurately than in single steps. Second, the learning rate $\cone$ is modulated: the increased length of the evolution path 
as computed in \eqref{eq-path-effect} acts in effect similar to an increased learning rate by a factor of up to $\cc^{-1/2}$.

As a last step, we combine \eqref{eqrankmu} and \eqref{eqrankone}.

\subsection{Combining Rank-$\mu$-Update and Cumulation \label{sec:combine}}
The final CMA update of the covariance matrix combines
\eqref{eqrankmu} and \eqref{eqrankone}.
\equationframe{{
\begin{eqnarray}
  \Cgg &=& (1\underbrace{
              \!{}-\cone-\cmu\sumw
              }_{\hspace*{-9em}\text{can be close or equal to $0$}\hspace*{-9em}}
              )\,\Cg 
             \nonumber\\ && 
       \qquad 
            + \left.\cone\,
             \underbrace{\pcgg{\pcgg}^\T
           \hspace{-1ex}}_{\!\!\mbox{rank-one update}\!\!} 
             \;\right. 
    \left. 
       +\;\cmu
       \underbrace{\summ{i}{\lam}w_i \,\ygg_\ilam \rklam{\ygg_\ilam}^\T
       \hspace{-1ex}}_{\mbox{rank-$\mu$ update}}
       \right.\label{eqcov}
\end{eqnarray}
}}
where
\where{$\cone\approx 2/n^2$. }
\where{$\cmu\approx{\min(\mueff/{\N^2}, 1-\cone})$. }
\where{$\ygg_\ilam = (\xgg_{i:\lam}-\xmeang)/\sigg$. }
\where{$\sumw = \summ{i}{\lam}w_i \approx{-\cone}/{\cmu}$, but see also \eqsref{eq-def-w} and\eqrefadd{eq-def-w-dyn} in Appendix\nref{sec:algorithm}.}
\whereende
\Eqref{eqcov} reduces to \eqref{eqrankmu} for $\cone=0$ and to
\eqref{eqrankone} for $\cmu=0$. The equation combines the advantages of
\eqsref{eqrankmu} and \eqrefadd{eqrankone}. On the one hand, the information
from the entire population is used efficiently by the
so-called rank-$\mu$ update.  On the other hand,
information of correlations \emph{between} generations is exploited by using
the evolution path for the rank-one update.
The former is important in large populations, the latter is 
particularly important in small populations.

\section{Step-Size Control\index{step-size control}\label{secsscontrol}} 
\begin{figure}[tb]
  \begin{center}
    \begin{minipage}{\textwidth}
      \includegraphics[width=0.99\textwidth]{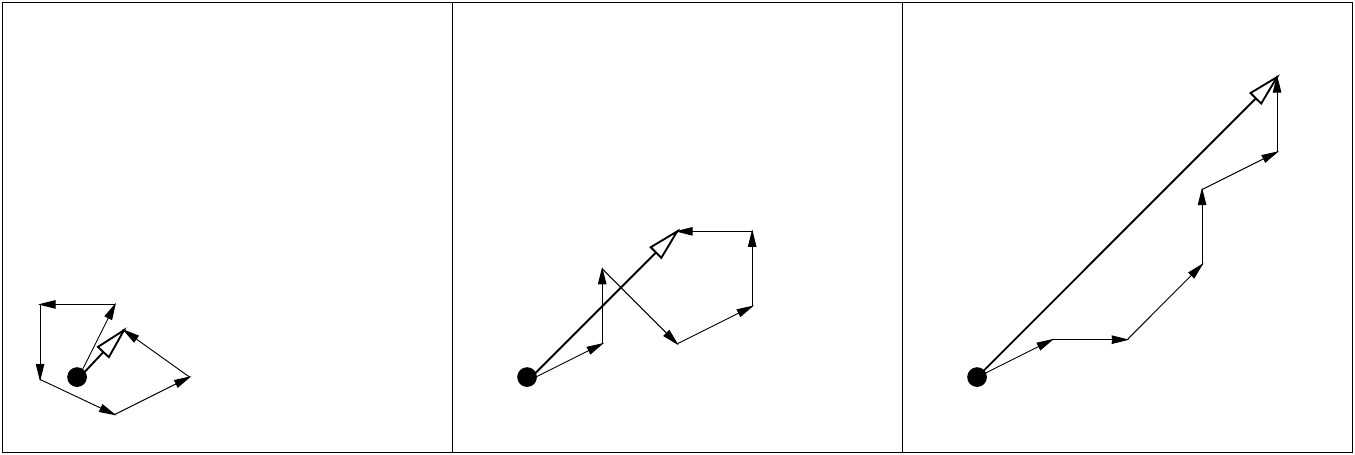}
    \end{minipage}
  \end{center}
  \caption[Pfade]{\label{figpfade}Three evolution paths of
  respectively six steps from different selection situations
  (idealized). The lengths of the \emph{single} steps are all
  comparable. The length of the evolution paths (sum of steps) is
  remarkably different and is exploited for step-size control}
\end{figure}

The covariance matrix adaptation, discussed in the last section, does
not explicitly control the ``overall scale'' of the distribution, the
step-size. The covariance matrix adaptation increases or decreases 
the scale only \emph{in a single direction} for each selected step---or it decreases the scale
by fading out old information by a given, non-adaptive factor. 
Less informally, we have two specific reasons to
introduce a step-size control in addition to the adaptation rule
\eqref{eqcov} for $\Cg$.
\begin{enumerate}
  \item The \emph{optimal} overall step length cannot be well approximated by
    \eqref{eqcov}, in particular if $\mueff$ is chosen larger than
    one.
\begin{einschub}%
  For example, on $\fsphere(\ve{x})=\summ{i}{\N}x_i^2$, given $\Cg=\Id$ and
  $\lam\le\N$, 
  the optimal step-size $\sig$ equals approximately
  $\mu\,{\sqrt{\fsphere(\ve{x})}}/{\N}$ with equal recombination weights 
  \cite{Beyer:2001b,Rechenberg:1994} and $1.25\,\mueff{\sqrt{\fsphere(\ve{x})}}/{\N}$ 
  with optimal recombination weights \cite{arnold2006weighted}.\footnote{%
  Because recombination then reduces the size of the realized step by a factor of $\sqrt{\mu}$ or $\sqrt{\mueff}$ (under random selection or in large dimension), the effective optimal steps are proportional to $\sqrt{\mu}$ or $1.25\sqrt{\mueff}$, respectively.
  }\kom{

mueff = (sum |wi|)^2 / sum wi^2 
      = (sum |wi|)^2 / (lam * W_lam)
mueff(Ei) = (sum |Ei|)^2 / sum Ei^2
      = (sum |Ei|)^2 / (lam * W_lam)
      = (sum |Ei|)^2 / lam
      = (lam * 0.79)^2 / lam 
      = 0.79^2 lam 
      = 0.64 lam
 
W_lam := mean wi^2
  W_lam = (sum |wi|)^2 / (mueff * lam) \propto sum |wi|
if sum |wi| = 1 then 
  W_lam is tiny (W_lam = 1 / (mueff * lam))
if wi = Ei then 
  W_lam = 1 and mueff = lam^2 0.798^2 / lam 
                      = lam * 0.798^2
                      = lam * 0.64
Arnold (wi=Ei): 
  wi_opt* = Ei
  sigma_opt* = 1

re-normalized (sum |wi| = 1):
  wi_opt* = Ei / sum |Ei|
  sigma_opt* = sum |E_i| = lam * 0.798 = mueff / 0.798 
             = 1.25 * mueff

} This
  dependency on $\mu$ or \mueff\ can not be realized by \eqref{eqrankmu} or \eqref{eqcov}.
\end{einschub}
\item The largest reliable learning rate for the covariance matrix
  update in \eqref{eqcov} is too slow to achieve competitive change
  rates for the overall step length. 
  \begin{einschub}%
    To achieve optimal performance on $\fsphere$ with an Evolution
    Strategy with weighted recombination, the overall step length 
    must decrease by a factor of about
    $\exp(0.25)\approx1.28$ within $\N$ function
    evaluations, as can be derived from progress formulas as in 
    \cite{arnold2006weighted} and
      \cite[p.\,229]{Beyer:2001b}.  That is, the time horizon for the
    step length change must be proportional to $\N$ or shorter.  From
    the learning rates \cone\ and \cmu\ in \eqref{eqcov} follows that the
    adaptation is too slow to perform competitive on $\fsphere$
    whenever $\mueff\ll\N$. This can be validated by simulations even
    for moderate dimensions, $\N\ge10$, and small $\mueff\le1+\ln\N$.
  \end{einschub}

\end{enumerate}
To control the step-size $\sigg$ we utilize an evolution
path\index{evolution path}, \ie\ a sum of successive steps (see also
Sect\pref{sec:cumulation}). The method can be applied 
independently of the
covariance matrix update and is denoted as \emphindex{cumulative path
  length control}, {cumulative step-size control}, or
\auge{cumulative step length adaptation (CSA)}. The length of an
evolution path\kom{ (compare Sect\pref{sec:cumulation})} is exploited, based
on the following reasoning, as depicted in Fig\pref{figpfade}.
\begin{itemize}
\item Whenever the evolution path is short, single steps cancel each other
  out (Fig\pref{figpfade}, left).  Loosely speaking, they are
  anti-correlated.  If steps extinguish each other, the step-size
  should be decreased.
\item Whenever the evolution path is long, the single steps are
  pointing to similar directions (Fig\pref{figpfade}, right).  Loosely
  speaking, they are correlated.  Because the steps are similar, the
  same distance can be covered by fewer but longer steps into the same
  directions. In the limit case, when consecutive steps have
  identical direction, they can be replaced by any of the enlarged single
  step.  Consequently, the step-size should be increased.
\item In the desired situation the steps are (approximately)
  perpendicular in expectation and therefore uncorrelated
  (Fig\pref{figpfade}, middle).
\end{itemize}
To decide whether the evolution path is ``long'' or ``short'', we
compare the length of the path with its
expected length under random selection\footnote{%
  Random selection means that the index \mbox{$i:\lam$} (compare
  \eqref{eqreco}\kom{ and \eqrefadd{eqcov}}) is independent of the value of
  $\xgg_{i:\lam}$ for all $i=1,\dots,\lam$, e.g.\ ${i:\lam}=i$. },
where consecutive steps are independent and therefore
uncorrelated (uncorrelated steps are the
desired situation). If selection biases the evolution path to be
longer then expected, $\sig$ is increased, and, vice versa, if
selection biases the evolution path to be shorter than expected,
$\sig$ is decreased. In the ideal situation, selection does not
bias the length of the evolution path and the length equals its
expected length under random selection.

 In practice, to construct the evolution path, $\ps$, the same
 techniques as in \eqref{eqpc} are applied.  In contrast to
 \eqref{eqpc}, a \emph{conjugate} evolution path\index{conjugate
 evolution path} is constructed, because the expected length of the
 evolution path $\pc$ from \eqref{eqpc} depends on its direction
 (compare \eqref{eqpcdist}). Initialized with $\ps^{(0)}=\ve{0}$, the
 conjugate evolution path reads
\equationframe{{\vspace{-1.5ex}
\begin{eqnarray}\label{eqcumsig}
  \psgg &=& (1-\cs)\psg + \sqrt{\cs(2-\cs)\mueff}\;
  {\Cg}^{-\frac{1}{2}} \;\frac{\xmeangg-\xmeang}{\cm\sigg} 
\end{eqnarray}
\vspace{-2ex}
}}
where 
\where{$\psg\in\Rn$ is the conjugate evolution path at generation
  $g$.}
\where{$\cs<1$.  Again, ${1}/{\cs}$ is the backward time
  horizon of the evolution path (compare \eqref{eqcovtime}). For small
  $\mueff$, a time horizon between $\sqrt{\N}$ and $\N$ is reasonable.
}
\where{$\sqrt{\cs(2-\cs)\mueff}$ is a normalization constant, see
  \eqref{eqpc}. }
\where{${\Cg}^{-\frac{1}{2}} \stackrel{\mathrm{def}}{=} \Bg{\Dg}^{-1}{\Bg}^\T$, where
  $\Cg=\Bg\rklam{\Dg}^{2}{\Bg}^\T$ is an eigendecomposition of $\Cg$, where
  ${\Bg}\kom{^\T\Bg=\Bg{\Bg}^\T=\Id}$ is an orthonormal basis of
  eigenvectors, and the diagonal elements of the diagonal matrix $\Dg$
  are square roots of the corresponding positive eigenvalues\tutorial{
  (cf.\ Sect\pref{sec:eigen})}. }
\whereende
For $\Cg=\Id$, we have $\invsqrtC{g}=\Id$ and \eqref{eqcumsig}\kom{
for $\psgg$} replicates
\eqref{eqpc}\kom{ for $\pcgg$}. The transformation ${\Cg}^{-\frac{1}{2}}$ re-scales
the step $\xmeangg-\xmeang$ within the coordinate system given by
$\Bg$.
\begin{einschub}%
  The single factors of the transformation
  ${\Cg}^{-\frac{1}{2}}{=\Bg{\Dg}^{-1}{\Bg}^\T}$ can be explained as
  follows (from right to left):

\begin{description}
\item ${\Bg}^\T$ rotates the space such that the columns of $\Bg$,
  \ie\ the principal axes of the distribution $\textNormal{\ve{0},\Cg}$,
  rotate into the coordinate axes. Elements of the resulting vector
  relate to projections onto the corresponding eigenvectors.

\item ${\Dg}^{-1}$ applies a (re-)scaling such that all axes become
equally sized.

\item $\Bg$ rotates the result back into the original coordinate
system. This last transformation ensures that the principal axes of
the distribution are not rotated by the overall transformation and
directions of consecutive steps are comparable.
\end{description}
\end{einschub}%
Consequently, the transformation ${\Cg}^{-\frac{1}{2}}$ makes the
expected length of $\psgg$ independent of its direction, and for any
sequence of realized covariance matrices $\Cg_{g=0,1,2,\dots}$ we have
under random selection $\psgg\sim\NormalNullI$, given
$\ps^{(0)}\sim\NormalNullI$ \cite{Hansen:1998}.

To update $\sigg$, we ``compare'' $\|\psgg\|$ with its expected length
$\chiN$, that is
\begin{equation}\label{eqlogsig}
  \ln\siggg = \ln\sigg +
  \frac{\cs}{\ds}\left(\frac{\|\psgg\|}{\chiN} - 1\right) \enspace,
\end{equation}
where
\where{$\ds\approx1$, damping parameter, scales the change magnitude
  of $\ln\sigg$.  The factor $\cs/\ds/\chiN$ is based on in-depth
  investigations of the algorithm \cite{Hansen:1998}. }
\where{$\chiN= \sqrt{2}\, \mathrm{\Gamma}(\frac{n+1}{2})
  /\mathrm{\Gamma}(\frac{n}{2})\approx\sqrt{\N}+{\cal O}(1/\N)$,
  expectation of the Euclidean norm of a $\NormalNullI$ distributed
  random vector. } \whereende
For $\|\psgg\|=\chiN$ the second summand in \eqref{eqlogsig} is zero,
and $\sigg$ is unchanged, while $\sigg$ is increased for
$\|\psgg\|>\chiN$, and $\sigg$ is decreased for
$\|\psgg\|<\chiN$.
\begin{einschub}Alternatively, we might use the squared norm $\|\psgg\|^2$ in
\eqref{eqlogsig} and compare with its expected
value $\N$ \cite{Arnold:2004}. In this case \eqref{eqlogsig} would
read
\begin{eqnarray}
  \ln\siggg 
	&=& \ln\sigg + \frac{\cs}{2\ds}\rklam{\frac{\|\psgg\|^2}{\N}-1}
         \label{eqlogsigsqr} \enspace.
\end{eqnarray}
This update performs rather similar to \eqref{eqlogsig}, while it presumable leads to  faster step-size increments and slower step-size decrements. 
\end{einschub}

 The step-size change is unbiased on the log scale,
because $\Extext{\ln\siggg\left|\sigg\right.} = \ln\sigg$ for
$\psgg\sim\NormalNullI$. The role of unbiasedness is discussed in
Sect\pref{stationarity}. \Eqsref{eqcumsig} and \eqrefadd{eqlogsig}
cause successive steps of the distribution mean $\xmeang$ to be
approximately $\invC{g}$-conjugate.
\begin{einschub}
\begin{sloppypar}\noindent
  In order to show that successive steps are approximately $\invC{g}$-conjugate
  first we remark that \eqsref{eqcumsig} and \eqrefadd{eqlogsig} adapt
  $\sig$ such that the length of $\psgg$ equals approximately $\chiN$.
  Starting from $(\chiN)^2 \approx {\|{\psgg}\|}^2=
  {{\psgg^{\T}\psgg}}={\mathrm{RHS}^\T\mathrm{RHS}}$ of
  \eqref{eqcumsig} and assuming that the expected squared \emph{length} of
  $\invsqrtC{g}(\xmeangg-\xmeang)$ is unchanged by selection (unlike its direction) we get
\end{sloppypar}
\begin{equation}
  \label{eq:pspercm}
   {\psg^{\T}\invsqrtC{g}(\xmeangg-\xmeang)}\approx0\enspace, 
\end{equation}
and\kom{ transforming
$\psg$ into the given coordinate system yields}
\begin{equation}
  \label{eq:ppercm}
  {\rklam{\sqrtC{g}\psg}^\T\invC{g}\rklam{\xmeangg-\xmeang}}\approx 0 \enspace. 
\end{equation}%
Given $1/(\cone+\cmu)\gg1$ and \eqref{eq:pspercm} we assume also
${\psgminone^{\T}\invsqrtC{g}(\xmeangg-\xmeang)}\approx0$ and derive
  \begin{equation}
      {\rklam{\xmeang-\xmean^{(g-1)}}^\T
      {{\Cg}^{-1}\rklam{\xmeangg-\xmeang}}} \approx 0 \enspace.
  \end{equation}
  That is, the steps taken by the
  \bookchapter{distribution} mean become
  approximately ${\Cg}^{-1}$-conjugate.
\end{einschub}
Because $\sigg>0$, \eqref{eqlogsig} is equivalent to 
\equationframe{{
\begin{equation}\label{eqsig}
{\siggg = \sigg
  \exp\rklam{\frac{\cs}{\ds}\rklam{\frac{\|\psgg\|}{\chiN}-1}}} 
\end{equation}
}}
\kom{The length of the evolution path is an intuitive and
  empirically well validated concept for an approximate optimality
  measure for the overall step length.  For $\mueff>1$ it is the best
  measure to our knowledge. Nevertheless, it fails to adapt nearly
  optimal step-sizes on very noisy objective functions
  \cite{Beyer:2003}.  }

\bookchapter{The length of the evolution path is an intuitive and
  empirically well validated goodness measure for the overall step
  length.  For $\mueff>1$ it is the best measure to our knowledge.\footnote{%
Recently, two-point adaptation has shown to achieve 
similar performance \cite{hansen2014assess}. 
}
  Nevertheless, it fails to adapt nearly optimal step-sizes on very
  noisy objective functions \cite{Beyer:2003}.  }

\newcommand{\singleruncaption}{Above: function value (\textit{thick
    line}), $\sigma$ (lower graph), $\sqrt{\mathtt{cond}(\ma{C})}
  $ 
  (upper graph). Middle: $\sqrt{\mathtt{diag}(\ma{C})}$, index
  annotated.  Below: square root of the eigenvalues of $\ma{C}$, \ie\ 
  $\mathtt{diag}(\ma{D})=[d_{11},\dots,d_{\N\N}]$, versus number of
  function evaluations.}
\newcommand{\figuresingleruns}[4]{
\begin{figure}
  \begin{center}
    \begin{minipage}{0.94\textwidth}
\frame{\parbox[t]{0.49\textwidth}{
      \includegraphics[width=0.496\textwidth,trim=0mm 0mm -0mm 0mm]%
                                      {#2}%
      \\[-3ex]
      \centerline{\figpart{a} axis parallel #1}
}}
\frame{\parbox[t]{0.49\textwidth}{
      \includegraphics[width=0.496\textwidth,trim=-0mm 0mm 0mm 0mm]%
                                              {#3}%
      \\[-3ex]
      \centerline{\figpart{b} randomly oriented #1}
}}
#4
    \end{minipage}
  \end{center}
\end{figure}
} 
\newcommand{\figuresinglerunonsphere}[4]{
\begin{figure}
  \begin{center}
    \begin{minipage}{0.94\textwidth}
\frame{\parbox[t]{0.49\textwidth}{
      \includegraphics[width=0.496\textwidth,trim=0mm 0mm -0mm 0mm]%
                                      {#2}%
      \\[-3ex]
      \centerline{\figpart{a} #1, $\ma{B}^{(0)}=\Id$}
}}
\frame{\parbox[t]{0.49\textwidth}{
      \includegraphics[width=0.496\textwidth,trim=-0mm 0mm 0mm 0mm]%
                                              {#3}%
      \\[-3ex]
      \centerline{\figpart{b} #1, $\ma{B}^{(0)}$ randomly oriented}
}}
#4
    \end{minipage}
  \end{center}
\end{figure}
} 

\newcommand{\va}{\ve{a}} \newcommand{\vx}{\ve{x}}
\newcommand{\mA}{\ma{A}}
\section{Discussion\bookchapter{}\label{secdiscussion}}
The \cma\ is an attractive option for non-linear
  optimization, if ``classical'' search methods, e.g.\ quasi-Newton
  methods (BFGS) and/or conjugate gradient methods, fail due to a
  non-convex or rugged search landscape (e.g.\ sharp bends,
  discontinuities, outliers, noise, and local optima). Learning the
  covariance matrix in the \cma\ is analogous to learning the inverse
  Hessian matrix in a quasi-Newton method. In the end, any
  con\-vex\sstrich{quadratic} (ellipsoid) objective function is
  transformed into the spherical function $\fsphere$. This can reduce 
  the number of $f$-evaluations needed to reach a target $f$-value 
  on ill-conditioned and/or non-separable problems by
  orders of magnitude.

The \cma\ overcomes typical problems that are often
associated with evolutionary algorithms. 
\begin{enumerate}
  \item Poor performance on badly scaled and/or highly
    non-separable objective functions. \Eqref{eqcov} adapts
    the search distribution to badly scaled and non-separable
    problems. 
    
  \item The inherent need to use large population sizes. A
    typical, however intricate to diagnose reason for the failure of
    population based search algorithms is the degeneration of the
    population into a subspace.\footnote{%
    The same problem can be observed with the downhill simplex 
    method \cite{nelder1965simplex} in dimension, say, larger than ten. 
    }
     This is usually prevented by
    non-adaptive components in the algorithm and/or by a large
    population size (considerably larger than the problem dimension).
    In the CMA-ES, the population size can be freely chosen, because
    the learning rates \cone\ and \cmu\ in \eqref{eqcov}\kom{ and the step-size
      control}\kom{and the time constant $\frac{1}{\cs}$ in
      \eqref{algps} and \eqref{algsig}} prevent the degeneration even
    for small population sizes, e.g.\ $\lam=9$.  Small population
    sizes usually lead to faster convergence,\kom{ in terms of number
      of function evaluations,} large population sizes help to avoid
    local optima\kom{ and therefore can improve the global search
      behavior}.

  \item Premature convergence of the population. Step-size control in
    \eqref{eqsig} prevents the population to converge prematurely. It
    does not prevent the search to end up in a {local} optimum. 

\end{enumerate}%
Therefore, the CMA-ES is highly competitive on a
considerable number of test functions
\cite{Hansen:1998,Hansen:2004b,Hansen:2003,Hansen:1997,Hansen:2001}
and was
  successfully applied to many real world problems.\footnote{%
    The author stopped to maintain the growing list of (at the time 120) published references to applications in 2009.}

  Finally, we discuss a few \auge{basic design principles} that were applied
  in the previous sections. 
  
\paragraph{Change rates}\index{change rates}
We refer to a change rate as the expected parameter change \emph{per
  sampled search point}, given a certain selection situation.  To
achieve competitive performance on a wide range of objective
functions, the possible change rates of the adaptive parameters need
to be adjusted carefully.  The \cma\ separately controls change rates
for the mean value of the distribution, $\xmean$, the covariance
matrix, $\ma{C}$, and the step-size, $\sig$.

\begin{itemize}
\item The change rate for the mean value $\xmean$, relative to the 
given sample distribution, is determined by \cm, and by the parent number and the
  recombination weights. The larger ${\mueff}$, the smaller is the
  possible change rate of $\xmean$\kom{, see \eqref{eqmueff}}.\footnote{%
  Given $\lam\not\gg\N$, then the mean change per generation is roughly 
  proportional to $\sig/\sqrt{\mueff}$, while the optimal step-size $\sig$ is roughly proportional to \mueff. 
  Therefore, the net change \emph{with optimal step-size} is proportional to 
  $\sqrt{\mueff}$ per generation. 
  Now considering the effect on the resulting convergence rate, 
  a closer approximation of the gradient adds another
  factor of $\sqrt{\mueff}$, such that the generational progress rate is 
  proportional to \mueff. Given $\lam/\mueff\approx4$, we have the remarkable
  result that the convergence 
  rate \emph{per $f$-evaluation} is roughly independent of \lam.}
  Similar holds for most evolutionary algorithms.%

\item The change rate of the covariance matrix $\ma{C}$ is explicitly
  controlled by the learning rates \cone\ and \cmu\ and therefore detached from
  parent number and population size. The learning rate reflects the
  model complexity. In evolutionary algorithms, the explicit control
  of change rates of the covariances, independently of population
  size and mean change, is a rather unique
  feature.

\kom{M. Sebag and A. Ducoulombier. Extending population-based
incremental learning to continuous search spaces. In Th. Back,
G. Eiben, M. Schoenauer, and H.-P. Schwefel, editors, Proceedings of
the 5 th Conference on Parallel Problems Solving from Nature, pages
418--427. Springer Verlag, 1998.}

\item The change rate of the step-size $\sig$ is explicitly controlled by the
  damping parameter $\ds$ and is in particular independent from the
  change rate of $\ma{C}$.  The time constant $1/\cs\le\N$ ensures a
  sufficiently fast change of the overall step length in particular
  with small population sizes.
\end{itemize}

%
\paragraph{Invariance}
{Invariance properties}\index{invariance properties} of a search
algorithm denote identical behavior on a set, or a class of objective
functions. Invariance is an important property of the
CMA-ES.\footnote{Special acknowledgments to Iv\'{a}n
Santib\'{a}\={n}ez-Koref for pointing this out to
me.}\kom{For example, let $f:\Rn\to\R, \vx\mapsto f(\vx)$ and
$\vx^{(0)}=\in[0,1]^n$.  Then,
\emphindex{scale invariance} means identical search behavior on all
$f_\alpha: \vx\mapsto f(\alpha\vx),
\vx^{(0)}=\frac{1}{\alpha}\va$, for all $\alpha\not=0$. Invariances
are highly desirable: }
Translation invariance should be taken for granted in continuous
domain optimization. Translation invariance means that the search
behavior on the function $\ve{x}\mapsto f(\ve{x}+\ve{a})$,
$\ve{x}^{(0)} = \ve{b}-\ve{a}$, is independent of $\ve{a}\in\Rn$.
Further invariances, e.g.\ invariance to certain linear transformations of the
search space, are highly desirable: they imply uniform performance on
classes of functions\footnote{%
However, most invariances are linked to a state space transformation. Therefore, uniform performance is only observed \emph{after} the state of the algorithm has been adapted.
}
and therefore allow for generalization of
empirical results.  In addition to translation invariance, the CMA-ES\tutorial{ exhibits} the following invariances.
\begin{itemize}
\item Invariance to order preserving (\ie\ strictly monotonic)
  transformations\index{order preserving
    transformations}\index{transformations!order preserving} of the
  objective function value. The \tutorial{algorithm only depends} on \emph{the ranking} of
  function values.
\item Invariance to angle preserving (rigid) transformations\index{angle
    preserving transformations}\index{ transformations!angle
    preserving} of the search space (rotation,
  reflection, and translation) if the initial search
  point\tutorial{ is} transformed accordingly.
\item Scale invariance\index{scale invariance} if the initial scaling,
  e.g.\ $\sig^{(0)}$, and the initial search
  point\tutorial{, $\xmean^{(0)}$,} are chosen
  accordingly.
\item Invariance to a scaling of variables\index{scaling of variables} (diagonal invariance)
  if the initial diagonal covariance matrix $\ma{C}^{(0)}$, and the initial
  search point, $\xmean^{(0)}$, are chosen accordingly.

\item Invariance to any invertible linear transformation of the
  search space, $\mA$, if the initial covariance matrix
  $\ma{C}^{(0)}=\mA^{-1}\rklam{\mA^{-1}}^\T$, and the initial
  search point\tutorial{, $\xmean^{(0)}$, are}
  transformed accordingly. Together with translation invariance, this can 
  also be referred to as \emphindex{affine invariance}, i.e.\ invariance to 
  affine search space transformations. 
\end{itemize}
Invariance should be a fundamental design criterion for any search
algorithm. Together with the ability to efficiently adapt the
invariance governing parameters, invariance is a key to
competitive performance.

\paragraph{Stationarity or Unbiasedness}\label{stationarity}\index{stationarity!design
  criterion}
An important design criterion for a \emph{randomized} search procedure
is \emphindex{unbiasedness}\index{unbiasedness!design criterion} of
variations of object and strategy parameters
\cite{Beyer:2001,Hansen:2001}\kom{not
  www-available:Hansen:1998,Beyer:2000}.  Consider random selection,
e.g.\ the objective function $f(\ve{x})= \mathtt{rand}$ to be
independent of $\ve{x}$. Then the population mean is unbiased if its
expected value remains unchanged in the next generation, that is
$\Extext{\xmeangg\left|\,\xmeang\right.}=\xmeang$. For the population
mean, stationarity under random selection is a rather intuitive
concept. In the CMA-ES, stationarity is respected for all parameters
that appear in the basic equation \eqref{eqmut}.  The distribution mean
$\xmean$, the covariance matrix $\ma{C}$, and $\ln\sig$ are unbiased.
Unbiasedness of $\ln\sig$ does not imply that $\sig$ is unbiased.
Under random selection,
$\Extext{\siggg\left|\,\sigg\right.}>\sigg$, compare
\eqref{eqlogsig}.\footnote{%
    Alternatively, if \eqref{eqsig} were designed to be unbiased
    for $\siggg$, this would imply that 
    $\Extext{\ln\siggg\left|\,\sigg\right.}<\ln\sigg$, in our opinion
    a less desirable alternative\kom{variant,option,version,possibility}. }

For distribution variances (or step-sizes) a bias toward increase or
decrease entails the risk of divergence or premature convergence,
respectively, whenever the selection pressure is low or when no 
improvements are observed. 
On noisy problems, a properly controlled bias towards increase can be
appropriate. It has the non-negligible disadvantage that the decision 
for termination becomes more difficult.

\kom
{
\section{Conclusion}
We have introduced the CMA Evolution Strategy

Reduced to estimating mean, step-size, and distribution shape. 
} 


\section*{Acknowledgments}
The author wishes to gratefully thank Anne Auger, Christian Igel,
Stefan Kern, and Fabrice Marchal for the many valuable comments on the
manuscript.


\bibliographystyle{plain}

\begin{appendix}
\newcommand{\Hsigg}{h_\sig^{(g+1)}}
\newcommand{\Hsig}{h_\sig}
\newpage
\section[Algorithm Summary: The CMA-ES]{\label{sec:algorithm}Algorithm Summary: The $(\mu/\mu_\mathrm{W},\lambda)$-\cma}
\newcommand{\ymean}{\overline{\ve{y}}}%
\renewcommand{\ymean}{\,{\langle{\ve{y}}\rangle}_\mathrm{w}}%
\newcommand{\zmean}{\,{\langle{\ve{z}}\rangle}_\mathrm{w}}%
\newcommand{\B}{\ma{B}}%
\newcommand{\C}{\ensuremath{\ma{C}}}%
\newcommand{\D}{\ma{D}}%
\newcommand{\alphacov}{\ensuremath{\alpha_\mathrm{cov}}}
\newcommand{\wa}{\ensuremath{w'}}%
\newcommand{\wb}{\ensuremath{w'}}%
\newcommand{\wia}{\ensuremath{w_i'}}%
\newcommand{\wib}{\ensuremath{w_i'}}%
\newcommand{\wja}{\ensuremath{w_j'}}%
\newcommand{\wjb}{\ensuremath{w_j'}}%
\newcommand{\w}{\ensuremath{w^{\circ}}}%
\newcommand{\mueffminus}{\ensuremath{\mueff^-}}%
\noindent Figure\nref{fig:algorithm} outlines the complete algorithm\footnote{%
With negative recombination weights in the covariance matrix, chosen here by default, the algorithm is sometimes denoted as aCMA-ES for active CMA \cite{jastrebski2006improving}.},  
  summarizing \eqsref{eqmut}, \eqrefadd{eqrecofinal}, \eqrefadd{eqpc},
  \eqrefadd{eqcov}, \eqrefadd{eqcumsig}, and \eqrefadd{eqsig}.
Used symbols, in order of appearance, are:

\begin{figure}
  \centering
\fbox{
\fbox{\parbox{\textwidth}{
\auge{Set parameters} 
\begin{itemize}
\item[] Set parameters $\lam$, $w_{i=1\dots\lam}$, $\cm$, $\cs$, {$\ds$},
  {$\cc$}, $\cone$, and {$\cmu$} according
  to Table\nref{tabdefpara}.
\end{itemize}

\auge{Initialization}
\begin{itemize}
\item[] Set evolution paths $\ps=\ve{0}$,
    $\pc=\ve{0}$, covariance matrix $\ma{C}=\Id$, and $g=0$.

\item[] Choose distribution mean $\xmean\in\Rn$ and step-size
  $\sig\in\R_{>0}$ 
  problem dependent.$^1$
\end{itemize}

\auge{Until termination criterion met}, $g\gets g+1$

~\\
\hspace*{\fill}\parbox{0.95\textwidth}{
Sample new population of search points, for $k=1,\dots,\lam$
\begin{eqnarray}
   \z_k &\sim&\NormalNullI \\
   \y_k &=& \B\D\z_k \quad\sim \Normal{\ve{0},\C}\enspace  \label{algsampley}\\
   \x_{k} &=& \xmean + \sig\y_k \quad\sim\Normal{\xmean,\sig^2\C}\enspace
   \label{algwgen}
\end{eqnarray}
Selection and recombination
\begin{eqnarray}
   \ymean &=& \summ{i}{\mu}w_i\,\y_\ilam
   \quad\mbox{ where }\summ{i}{\mu}w_i=1,~\text{$w_i>0$ for $i=1\dots\mu$}\\
    \xmean &\gets& \xmean + \cm\sig\ymean 
      \qquad\text{equals $\summ{i}{\mu}w_i\,\x_\ilam$ if $\cm=1$} \label{algreco}
\end{eqnarray}

Step-size control
\begin{eqnarray}
  \ps &\gets& (1-\cs)\ps 
        + \sqrt{\cs(2-\cs)\mueff}\;
  {\C}^{-\frac{1}{2}} \ymean\label{algps}\\
   \sig &\gets& \sig\times
      \exp\rklam{\frac{\cs}{\ds}\rklam{\frac{\|\ps\|}{\chiN}-1}}\label{algsig}
\end{eqnarray}

Covariance matrix adaptation
\begin{eqnarray}
  \pc &\gets& (1-\cc)\pc + \Hsig\sqrt{\cc(2-\cc)\mueff} 
     \ymean\label{algpc}\\
\w_i &=& \label{eq-def-w-dyn}
\text{$w_i\times(1$ if $w_i\ge0$ else $\displaystyle{\N}/{\|\C^{-\frac{1}{2}}\y_\ilam\|^2})$}
\\
  \C &\gets& (1+\underbrace{\cone\delta(\Hsig)-\cone-\cmu
              \sumw}_{\text{usually equals to $0$ }}
              )\,\C + \cone
    \pc\pc^\T 
    + \;\cmu
       \summ{i}{\lam}\w_i \,\y_\ilam \y_\ilam^\T\label{algcov}
\end{eqnarray}
\\[-4ex]
\rule{5em}{0.1ex}\\
{\hspace*{1.5em}$^1$\footnotesize
  The optimum should presumably be within the initial cube $ \xmean\pm
  3\sig(1,\dots,1)^\T$. If the optimum is expected to be in the
  initial search interval $[a,b]^\N$ we may choose the initial search
  point, $\xmean$, uniformly randomly in $[a,b]^\N$, and
  $\sig=0.3(b-a)$.  Different search intervals $\mathrm{\Delta} s_i$
  for different variables can be reflected by a different
  initialization of $\ma{C}$, in that the diagonal elements of
  $\ma{C}$ obey $c_{ii}=(\rmDelta s_i)^2$. However, the $\rmDelta
  s_i$ should not disagree by several orders of magnitude. Otherwise a
  scaling of the variables should be applied. }

}}}}

  \caption[CMA]{\label{fig:algorithm}The $(\mu/\mu_\mathrm{W},\lambda)$-CMA 
    Evolution Strategy. Symbols: see text}

\end{figure}%
\where{$\y_k\sim\Normal{\ve{0},\ma{C}}$, for $k=1,\dots,\lam$, are
   realizations from a multivariate normal distribution with
   zero mean and covariance matrix $\ma{C}$.  }
\where{$\B,\D$ result from an eigendecomposition of the
   covariance matrix $\C$ with
   $\C=\B{\D}^2{\B}^\T=\B{\D}{\D}{\B}^\T$\tutorial{ (cf.\
   Sect\pref{sec:eigen})}.  Columns of $\B$ are an orthonormal basis
   of eigenvectors.  Diagonal elements of the diagonal matrix $\D$ are
   square roots of the corresponding positive eigenvalues. While
   \eqref{algsampley} can certainly be implemented using a Cholesky
   decomposition of \C, the eigendecomposition is needed to correctly 
   compute $\C^{-\frac{1}{2}} = \B{\D}^{-1}{\B}^\T$ for \eqsref{algps} 
   and \eqrefadd{eq-def-w-dyn}.  }
\where{$\x_k\in\Rn$, for $k=1,\dots,\lam$. Sample of $\lam$ search
   points. }
\where{$\ymean = \summ{i}{\mu}w_i\,\y_\ilam$, step of the
  distribution mean disregarding step-size $\sig$. }
\where{$\y_\ilam = (\x_\ilam-\xmean)/\sig$, see $\x_\ilam$ below.}
\where{$\x_\ilam\in\Rn$, $i$-th best point out of
  $\x_1,\dots,\x_\lam$ from \eqref{algwgen}. The index
  \mbox{$\ilam$} denotes the index of the $i$-th ranked point, that
  is $f(\x_{1:\lam})\le f(\x_{2:\lam})\le\dots\le
  f(\x_{\lam:\lam})$.}
\kom{\where{$\ps\in\Rn$ evolution  path for $\sig$. }}%
\where{$
\mu 
    = \left|\{w_i\,|\, w_i > 0\}\right|
    = \summ{i}{\lam} \mathbf{1}_{(0, \inf)}(w_i) \ge 1$ 
  is the number of strictly positive recombination weights.
}
\where{$\mueff = \rklam{\summ{i}{\mu}w_i^2}^{-1}$ is the variance
  effective selection mass, see \eqref{eqmueff}. Because $\summ{i}{\mu}|w_i| = 1$, we have
  $1\le\mueff\le\mu$.}
\where{${\C}^{-\frac{1}{2}} \stackrel{\mathrm{def}}{=} \B{\D}^{-1}{\B}^\T$,
  see $\B,\D$ above. The matrix $\D$ can be inverted by inverting
  its diagonal elements.  From the definitions we find that
  ${\C}^{-\frac{1}{2}}\y_i = \B\z_i$, and
  ${\C}^{-\frac{1}{2}}\!\ymean = \B 
  \summ{i}{\mu}w_i\,\z_\ilam$.  }
\where{$\chiN=\sqrt{2}\,
  \mathrm{\Gamma}(\frac{n+1}{2})/\mathrm{\Gamma}(\frac{n}{2})
  \approx\sqrt{\N}\rklam{1-\frac{1}{4\N}+\frac{1}{21\N^2}}$.}
\kom{\where{$\pcgg\in\Rn$, evolution path for $\ma{C}$. }}
\where{$\Hsig=\left\{
    \begin{array}{ll}
      1 &\mbox{if }\frac{\|\ps\|}{\sqrt{1-(1-\cs)^{2(g+1)}}} < (1.4 + \frac{2}{\N+1})\chiN\\
      0 &\mbox{otherwise}
    \end{array}\right.$, where $g$ is the generation number. 
  The Heaviside function $\Hsig$ stalls the update of $\pc$ in
  \eqref{algpc} if $\|\ps\|$ is large. This prevents a too fast
  increase of axes of $\C$ in a linear surrounding, \ie\ when the
  step-size is far too small. This is useful when the initial
  step-size is chosen far too small or when the objective function
  changes in time. }
\kom{\where{$\Cg$, symmetric and positive definite $\N\times\N$ covariance
  matrix. } }
\where{$\delta(\Hsig)=(1-\Hsig)\cc(2-\cc)\le 1$ is of minor
  relevance. In the (unusual) case of $\Hsig=0$, it substitutes for
  the second summand from \eqref{algpc} in \eqref{algcov}.}
\where{$\sumw=\sum_{i=1}^\lam w_i$ is the sum of the recombination weights, see \eqref{eq-w-a}--\eqref{eq-def-w}. We have $-\cone/\cmu \le\sumw\le 1$ and for the default population size $\lam$, we meet the lower bound $\cmu\sumw=-\cone$.
%
}
\whereende
\newpage

\newcommand{\alphamueffminus}{\ensuremath{\alpha_{\mueff}^-}}
\newcommand{\alphaposdef}{\ensuremath{\alpha_\mathrm{pos\,def}^-}}
\paragraph{Default Parameters}
The (external) strategy parameters are $\lam$, $w_{i=1\dots\lam}$, $\cm$,
$\cs$, {$\ds$}, {$\cc$}, $\cone$, and {$\cmu$}. {Default
strategy parameter values} are given in Table\nref{tabdefpara}.
\begin{table}
\caption{\label{tabdefpara}Default Parameters (in 2016), where 
  $\mu=|\{w_i>0\}|=\lfloor\lam/2\rfloor$,
  $\mueff=\frac{(\summ{i}{\mu}\wia)^2}{\summ{i}{\mu}\wia^2}\in[1, \mu]$, 
  $\mueffminus=\frac{(\sum_{i=\mu+1}^{\lam}\wia)^2}{\sum_{i=\mu+1}^{\lam}\wia^2}$,
  $\summ{i}{\mu}w_i=1$, and $\possum{w_j}$ is the sum
  of all positive, and $-\negsum{w_j}$ the sum of all negative $w_j$-values, 
  i.e., $\alphamuminus=\negsum{\wb_j}\ge0$.
  Apart from $w_i$ for $i>\mu$, all parameters are taken from \cite{Hansen:2009} 
  with only minor modifications
}

\fbox{\parbox{\textwidth}{
Selection and Recombination:
\begin{eqnarray}\label{eq-def-lam}
\lam&=&4+\lfloor3\,\ln\N\rfloor \qquad\text{can be increased}
\end{eqnarray}\vspace{-3ex}
\begin{eqnarray}
\wia &=& \ln\displaystyle\frac{\lam+1}{2}-\ln i \quad\text{for $i=1,\dots,\lambda$}\qquad\text{preliminary convex shape} \label{eq-w-a}
\end{eqnarray}
\begin{align}\label{eq-alphamuminus}
  \alphamuminus & = 
					1 + \cone/\cmu
					\qquad\parbox{0.5\textwidth}{let $\cone + \cmu\sum w_i=\cone + \cmu - \cmu\negsum{w_i} $ be $0$}
\\
\label{eq-alphamueffminus}
  \alphamueffminus & = 1 + \frac{2\mueffminus}{\mueff+2}
  \qquad\parbox{0.5\textwidth}{bound \negsum{w_i} to be compliant with $\cmu(\mueff)$}
\\
\label{eq-alphaposdef}
  \alphaposdef & = \frac{1-\cone-\cmu}{\N\,\cmu
    } \qquad\parbox{0.45\textwidth}{\raggedright bound \negsum{w_i} to guaranty positive definiteness\hspace*{-2em}}
\end{align}
\begin{eqnarray}
w_i   &=& \label{eq-def-w}
\left\{\!\!
    \begin{array}{ll}
    \displaystyle\frac{1}{
    \possum{\wja}
    } \,\wia & \text{if $\wia\ge0$} \qquad\parbox{0.302\textwidth}{\raggedright 
    				positive weights sum to one} \\
    \displaystyle\frac{\min(\alphamuminus, \alphamueffminus, \alphaposdef)}{
    \negsum{\wja}
    }\,\wia 
		& \text{if $\wia<0$} \qquad\parbox{0.302\textwidth}{\raggedright 
					negative weights usually sum to $-\alphamuminus$}
    \end{array}\right. 
    %
    %
\end{eqnarray}
\begin{equation}
\cm = 1
\end{equation}

Step-size control:
\begin{equation}\label{eq-def-sig}
\begin{array}{l}
\displaystyle
\cs= \frac{\mueff+2}{n+\mueff+5}\\
\displaystyle
\ds = 1 + 2\,\max\left(0,\;\sqrt{\frac{\mueff-1}{n+1}}-1\right) + \cs
\end{array}
\end{equation}

Covariance matrix adaptation:
\begin{equation}
\cc=\frac{4 + \mueff/n}{n+4+2\mueff/n}
\label{eq-def-cc}
\end{equation}
\begin{equation}
\cone = \frac{\alphacov}{(n+1.3)^2 + \mueff}  
                    \quad \text{with}\;\alphacov = 2
\label{eq-def-cone}
\end{equation}
\begin{equation}\label{eq-def-cmu}
\cmu{}=
   \min\!\left(1-\cone,\alphacov\frac{
                                       1/4 + \mueff + 1/\mueff - 2
                    }{(n+2)^2+\alphacov\mueff/2}\right) 
                    \quad \text{with}\;\alphacov = 2
\end{equation}
}}
\end{table}%
An in-depth discussion of most parameters is given in
\cite{Hansen:2001}. 
  
  The given setting for the default \auge{negative weights} was introduced in 2016. 
  The setting is somewhere between uniform weights \cite{jastrebski2006improving} 
  and mirrors of the positive weight values \cite{arnold2006weighted,hansen2010benchmarking}. 
  The choice is a compromise between avoiding increasingly large negative values which lead to a large variance reduction in a single direction in \C\ while still giving emphasis on the selection \emph{differences} in particular for weights close to the median rank. We attempt to scale all negative weights such that the factor in front of \C\ in \eqref{algcov} becomes $1$. That is, we have by default no decay on \C\ and the variance added to the covariance matrix by the positive updates equals, in expectation, to the variance removed by the negative updates. 

\begin{einschub}%
  %
  %
%
  Specifically, we want to achieve $\cone+\cmu\sumw=0$, 
  that is 
  \begin{align*}
  \cone & = -\cmu\sumw \\ 
  \cone/\cmu &= -\left(\possum{w_j} - \negsum{w_j}\right) \\
  \cone/\cmu &= -1 + \negsum{w_j} \\
  1 + \cone/\cmu &= \negsum{w_j} \enspace,   
  \end{align*}
  hence the
  multiplier \alphamuminus\ in \eqref{eq-def-w} is set to $1 + \cone / \cmu$. 

  Choosing \negsum{w_j} in the order of $1$ is only viable if $\mueff\not\gg \mueffminus = {\left(\sum_{i=\mu+1}^\lam w_i\right)^2} / \;{\sum_{i=\mu+1}^\lam {w_i}^2}$, that is, if the variance effective update information from positive weights, \mueff, is not much larger than that from negative weights, \mueffminus. In the default setting, \mueffminus\ is about $1.2$ to $1.5$ times \emph{larger} than $\mueff$, because the curve $w_i$ versus $i$ flattens out for increasing $i$. In \eqref{eq-def-w} we use the bound $\alphamueffminus$, see \eqref{eq-alphamueffminus}, to (i) get a meaningful value for any choices of $\wia$, and (ii) preserve the effect from letting \cmu\ go to zero (eventually turning off the covariance matrix adaptation entirely).

  The apparent circular dependency between $w_i$, \alphamuminus, \cmu, \mueff, 
  and again $w_i$ can be resolved: 
  the variance effective selection mass $\mueff$ depends
  only on the \emph{relative} relation between the \emph{positive} weights, such that
  $\mueff(w_{1\dots\lambda}) = 
   \mueff(w_{1\dots\mu}) = 
   {(\summ{i}{\mu}w_i)^2}/{\summ{i}{\mu}w_i^2} = 
   \mueff(\wa_{1\dots\mu})$. That is, \mueff\ and \mueffminus\ 
   can be 
   computed already from $\wa_i$ of \eqref{eq-w-a}, from which \cmu\ can be computed, 
   from which \alphamuminus\ can be computed, from which the remaining negative weights
   $w_i$ can be computed. 
\end{einschub}   
  Finally, we also bound the negative weights via \eqref{eq-def-w}
  to guaranty positive definiteness of \C\ via \eqref{eq-def-w-dyn},
  thereby, possibly, re-introducing a decay on \C. 
  With the default setting for population size \lam\ and the default raw weight values, \alphaposdef\ in \Eqref{eq-def-w} leaves the weights unchanged. 
\begin{einschub}
\begin{sloppypar}\noindent
Specifically, to guaranty positive definiteness of the covariance matrix, we can bound the maximal variance subtracted in a single direction by the variance remaining after the decay on \C\ is applied in \eqref{algcov}, disregarding any added variance (worst case).
Defining $\negsum{w_i} = \sum_{i=\mu+1}^{\lam}|w_i|$ to be the sum of the absolute values of all negative weights, and assuming a (Mahalanobis-)variance of $\N$ from each negative summand of the weighted sum in \eqref{algcov}, we require 
\begin{equation}
\N\cmu\negsum{w_i} < 1-\cone-\sumw\cmu = 1-\cone-\cmu + \cmu\negsum{w_i}
\enspace.
\end{equation}
Solving for $\negsum{w_i}$ yields
\begin{equation}
\negsum{w_i} < \frac{1-\cone-\cmu}{(\N-1)\cmu}
\enspace.
\end{equation}
We use $\min(\dots, \frac{1-\cone-\cmu}{\N\cmu})$ as multiplier for setting $w_{i=\mu+1\dots\lam}$ in \eqref{eq-def-w} and normalize the variance from each respective summand $\y_\ilam\y_\ilam^\T$ via \eqref{eq-def-w-dyn} to $\N$, thereby bounding the variance reduction from negative weight values to the factor $\frac{\N-1}{\N}$. 
\end{sloppypar}
\end{einschub}
A Python implementation of the slightly intricate computation of the weights \eqref{eq-w-a}--\eqref{eq-def-w} can be found \href{https://gist.github.com/nikohansen/3eb4ef0790ff49276a7be3cdb46d84e9}{here}. 
 
The cumulation parameter $\cc$ for \eqref{eq-def-cc} is chosen $\propto 1/\N$. 
For learning a single direction in (almost) linear time, the setting $\cc \propto 1/\sqrt{\N}$ seems sufficient \cite{hansen2014principled}. 
Larger values are in general more robust to avoid feedback oscillations or instabilities and might allow for more freedom in the setting of \cone.
However, on the cigar function the relationship between \cc\ \emph{times dimension} ($\cc\times\N$) and the evaluations divided by dimension to reach a given target becomes perfectly invariant with increasing dimension.\footnote{
This invariance remains almost perfect when the learning rates \cone\ and \cmu\ are chosen $\propto\N^{-1.5}$ instead of $\propto\N^{-2}$ ($\propto\N^{-1.5}$ is however not a reliable setting). 
It changes to about $\cc\times\N^{1/2}$ when only the diagonal elements of the covariance matrix are adapted with a learning rate of $\propto1/\N$. }

\begin{figure}[tb]
 \vspace*{-0cm}
  \begin{center}
      \includegraphics[width=0.49\textwidth]{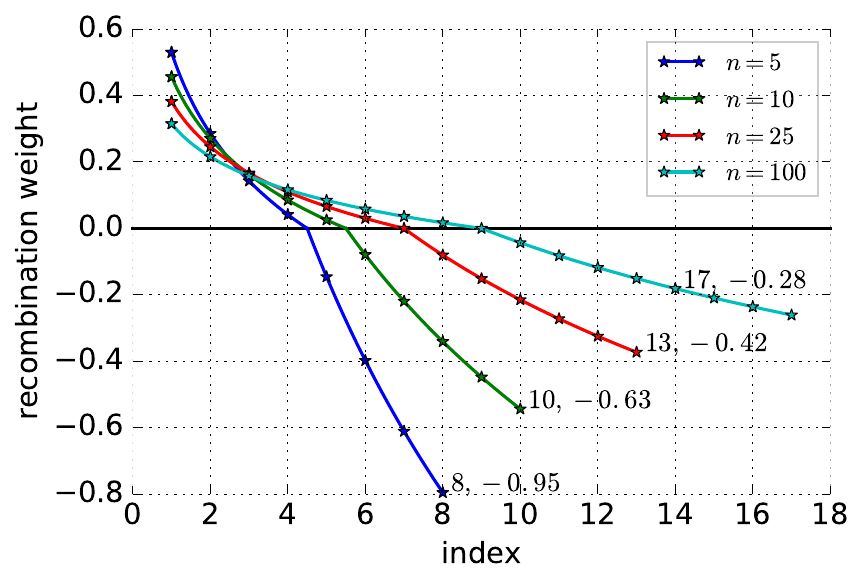}
            %
            %
      \includegraphics[width=0.49\textwidth]{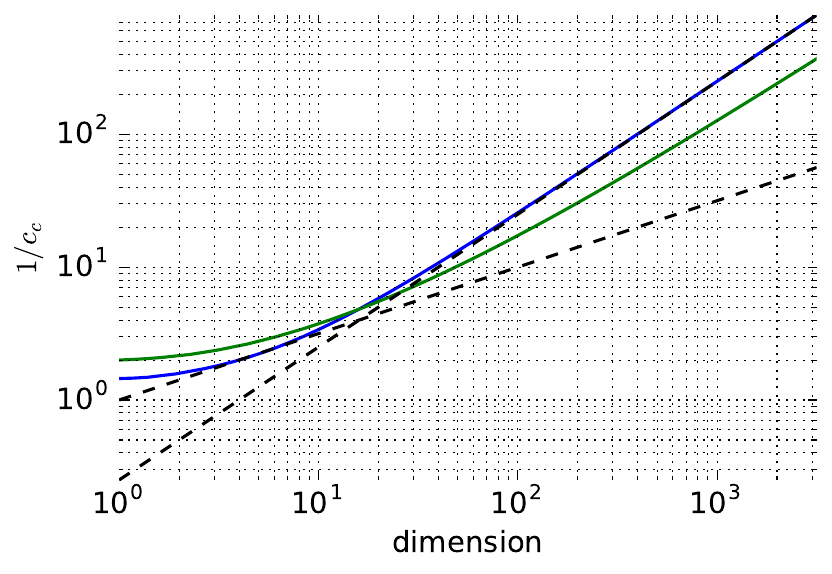}
            %
            %
 \vspace*{-0.3cm}
  \end{center}
  \caption[ccov]{\label{fig-cc-vs-dim} 
  Left: recombination weights from \eqref{eq-def-w} in Table~\ref{tabdefpara} for $\N=5, 10, 25, 100$. 
  Also given are the (default) population size \lam\ (corresponding to the last index) and the sum of weights. Positive weights sum to one. 
  Right: backward time horizon $1/\cc$ according to \eqref{eq-def-cc} (blue) and \eqref{eq-def-cc-gen} (green), and $\N/4$ and $\sqrt\N$ (both dashed) plotted against dimension \N. The blue line is above the green line by a factor of about $1.5, 2, 2.5=10/4$ for $\N=100, 1000, \infty$, respectively. }
\end{figure}

Adopting a technique from \cite[Eq.\ (11)]{suttorp2009ecm}, a generalized setting reads

\begin{equation}\label{eq-def-cc-gen}
\cc = \frac{\alpha_c(\N) + (\mueff/\N)^{\beta_c}}{\N^{\beta_c} + \alpha_c(\N) + 2\,(\mueff/\N)^{\beta_c}} \enspace.
\end{equation}
With $\alpha_c(\N)=10^{1-\N^{-1/3}}$ and $\beta_c=1$, the setting is slightly more robust in larger dimension but remains invariant, see Figure~\ref{fig-cc-vs-dim}.
When not all parameters of \C\ are subject to adaptation and $1/\cone = o(\N^{1.5})$, we suspect that $\beta_c < 1$ is the reasonable choice.
The last summand of the denominator of \eqref{eq-def-cc-gen} is chosen such that \cc\ approaches $1/2$ when 
$\mueff/\N$ approaches infinity. 

Similarly, the second summand of the denominator of \eqref{eq-def-cmu}, $\alphacov\mueff/2$, is chosen such that 
\cmu\ approaches one when $\alphacov\mueff$ approaches $2=2/(2-1)$ times the first summand.

Another technique, namely setting parameters depending on the degrees of freedom is developed in \cite{akimoto2020diagonal}.

The default parameters of
\eqsref{eq-def-w}--\eqrefadd{eq-def-cmu} are in particular chosen to be a
robust setting and therefore, to our experience, applicable to a wide
range of functions to be optimized. \emph{We do not recommend to change this
setting}, apart from increasing the population size $\lam$ in \eqref{eq-def-lam},\footnote{%
   Decreasing $\lam$ is not recommended. Too small values
   have strong adverse effects on the performance. }
 and possibly decreasing \alphacov\ on noisy functions.
 If the $\lam$-dependent default values for $w_i$ are used as given,
 the population size $\lam$ has a significant influence on the global
 search performance~\cite{Hansen:2004b}.  Increasing $\lam$ usually
 improves the global search capability and the robustness of the \cma,
 at the price of a reduced convergence speed. The convergence speed
 (per function evaluation) decreases at most linearly with $\lam$. Independent restarts with
 increasing population size \cite{Auger:2005a}, automated or manually
 conducted, are a useful policy to perform well on most problems.

\section{Implementational Concerns}
\newcommand{\offset}{f_{\mathrm{offset}}}
\newcommand{\alphaf}{\alpha_{f}}
\newcommand{\alphac}{\alpha_{c}}
We discuss a few implementational questions. 
  \subsection{Multivariate normal distribution} \begin{sloppypar} Let the
  vector $\z\sim\NormalNullI$ have independent, $(0,1)$-normally
  distributed components that can easily be sampled on a computer.  To
  generate a random vector $\y\sim\textNormal{\ve{0},\linebreak[0]\C}$
  for \eqref{algsampley}, we set $\y=\B\D\z$ (see above symbol
  descriptions of $\B$ and $\D$ and Sects\pref{sec:eigen} and
  \ref{sec:normal}, and compare lines 52--53 and 83--84 in the source
  code below).  Given $\y_k=\B\D\z_k$ and
  $\C^{-\frac{1}{2}}=\B\D^{-1}\B^\T$ we have
  $\C^{-\frac{1}{2}}\ymean=\B\summ{i}{\mu}w_i\,\z_\ilam$ (compare
  \eqref{algps} and lines 61 and 64 in the source code below).
 \end{sloppypar}
 
\subsection{Strategy internal numerical effort} In practice, the
  re-calculation of $\B$ and $\D$ needs to be done not until about
  $\max(1,\lfloor 1/(10\N(\cone+\cmu))\rfloor)$ generations. For reasonable
  $\cone+\cmu$ values, this reduces the numerical effort due to the
  eigendecomposition from ${\cal O}(\N^3)$ to ${\cal O}(\N^2)$ per
  generated search point, that is the effort of a matrix vector
  multiplication.

On a Pentium 4, 2.5 GHz processor
  the overall strategy internal time consumption is roughly
  $3\mal (\N+5)^2\mal 10^{-8}$ seconds per function evaluation
  \cite{Kern:2004}.

Remark that it is not sufficient to compute a Cholesky decomposition
of \C, because then \eqref{algps} cannot be computed correctly. 

\subsection{Termination criteria} 
  In general, the algorithm should be stopped whenever it becomes a
  waste of CPU-time to continue, and it would be better to restart
  (eventually with increased population size \cite{Auger:2005a}) or to
  reconsidering the encoding and/or objective function formulation. We
  recommend the following termination criteria \cite{Auger:2005a,Hansen:2009} that
  are mostly related to numerical stability:

\begin{itemize}
 \item \texttt{NoEffectAxis}: stop if adding a $0.1$-standard deviation
   vector in any principal axis direction of $\C$ does not change
   $\xmean$.\footnote{More formally, we terminate if $\xmean$ equals to
   $\xmean + 0.1 \, \sig d_{ii} \ve{b}_i$, where $i=(g\bmod
   \N)+1$, and $d_{ii}^2$ and $\ve{b}_i$ are respectively the $i$-th
   eigenvalue and eigenvector of $\C$, with $\|\ve{b}_i\|=1$.}

  \item \texttt{NoEffectCoord}: stop if adding $0.2$-standard
    deviations in any single coordinate does not change $\xmean$
    (\ie\ $m_i$ equals $m_i+0.2\,\sig c_{i,i}$ for any $i$).

  \item \texttt{ConditionCov}: stop if the condition number of the
    covariance matrix exceeds $10^{14}$.

  \item \texttt{EqualFunValues}: stop if the range of the best
    objective function values of the last $10+\lceil 30
    n/\lambda\rceil$ generations is zero.
    
  \item \texttt{Stagnation}: we track a history of the best and the median fitness in each iteration over the last 20\% but at least $120+30\N/\lambda$ and no more than $20\,000$ iterations. We stop, if in both histories the median of the last (most recent) 30\% values is not better than the median of the first 30\%. 

  \item \texttt{TolXUp}: stop if $\sig\times \max(\mathrm{diag}(\D))$
    increased by more than $10^4$. This usually indicates a far too
    small initial $\sig$, or divergent behavior.

\end{itemize}

  Two other useful termination criteria should be considered problem dependent:

\begin{itemize}
  
\item \texttt{TolFun}: stop if the range of the best objective
  function values of the last $10+\lceil 30 n/\lambda\rceil$
  generations and all function values of the recent generation is
  below \texttt{TolFun}. Choosing \texttt{TolFun} depends on the
  problem, while $10^{-12}$ is a conservative first guess.

\item \texttt{TolX}: stop if the standard deviation of the normal distribution is
  smaller than \texttt{TolX} in all coordinates and $\sig\pc$ is smaller than
  \texttt{TolX} in all components. By default we set \texttt{TolX} to
  $10^{-12}$ times the initial $\sigma$.

\end{itemize}

\subsection{Flat fitness}\index{flat fitness} In the case of equal function
  values for several individuals in the population, it is feasible to
  increase the step-size (see lines 92--96 in the source code
  below). This method can interfere with the termination criterion
  \texttt{TolFun}. In practice, observation of a flat fitness should
  be rather a termination criterion and consequently lead to a
  reconsideration of the objective function formulation.

\kom{\item Termination criteria: defaults, restart,...}

\subsection{Boundaries and Constraints} The handling of boundaries and 
 constraints is to a certain extend problem dependent. We discuss a few 
 principles and useful approaches. 

\begin{description}
 \item[Best solution strictly inside the feasible domain] If the optimal solution is \emph{not too close to the infeasible
 domain}, a simple and sufficient way to handle any type of boundaries and constraints is
 
 \begin{enumerate}
 \item setting the fitness as
 \begin{equation} \label{eq:pen}
   f_\mathrm{fitness}(\ve{x}) = f_\mathrm{max} +
   \|\ve{x}-\ve{x}_{\mathrm{feasible}}\| \enspace,
 \end{equation}
 where $f_\mathrm{max}$ is larger than the worst fitness in the feasible
 population or in the feasible domain (in case of minization) and $\ve{x}_{\mathrm{feasible}}$
 is a constant feasible point, preferably in the middle of the
 feasible domain.
 \item re-sampling any infeasible solution $\x$ until it
 become feasible. 
 \end{enumerate}

 \item[Repair available] as for example with box-constraints. 

\begin{description} 

  \item[Simple repair]  
 It is possible to simply repair infeasible individuals before the update equations are applied. This is not recommended, because the CMA-ES makes implicit assumptions on the distribution of solution points, which can be heavily violated by a repair. The main resulting problem might be divergence or too fast convergence of the step-size. However, a (re-)repair of changed or injected solutions for their use in the update seems to solve the problem of divergence \cite{Hansen:2011} (clipping the Mahalanobis distance of the step length to obey $\|\x-\xmean\|_{\sig^2\C}\le\sqrt\N + 2 \N / (\N + 2)$ seems to be sufficient). Note also that repair mechanisms might be intricate to implement, in particular if $\y$ or $\z$ are used for implementing the update equations in the original code. 

  \item[Penalization] We evaluate the objective function on a repaired
  search point, $\ve{x}_{\mathrm{repaired}}$, and add a penalty
  depending on the distance to the repaired solution.
\begin{equation} \label{eq:penbound}
  f_\mathrm{fitness}(\ve{x}) = f(\ve{x}_{\mathrm{repaired}}) +
  \alpha\, \|\ve{x}-\ve{x}_{\mathrm{repaired}}\|^2 \enspace.
\end{equation}
  The repaired solution is disregarded afterwards. 

  In case of box-boundaries, $\ve{x}_{\mathrm{repaired}}$ is set to
  the feasible solution with the smallest distance
  $\|\ve{x}-\ve{x}_{\mathrm{repaired}}\|$. In other words, components
  that are infeasible in $\ve{x}$ are set to the (closest) boundary
  value in $\ve{x}_{\mathrm{repaired}}$.  A similar boundary handling
  with a component-wise adaptive $\alpha$ is described in
  \cite{hansen:08}.
  
 \end{description}

  \item[No repair mechanism available] The fitness of the
  infeasible search point $\ve{x}$ might similarly compute to
  \begin{equation} \label{eq:pencon}
    f_\mathrm{fitness}(\ve{x}) = \offset
         + \alpha \sum_i {1\hspace{-0.33em}1}_{c_i>0}\times c_i(\ve{x})^2
%
   \end{equation}
 where, w.l.o.g., the (non-linear) constraints $c_i:\Rn\to\R,
 \ve{x}\mapsto c_i(\ve{x})$ are satisfied for $c_i(\ve{x})\le0$ , and
 the indicator function ${1\hspace{-0.33em}1}_{c_i>0}$ equals to one
 for $c_i(\ve{x})>0$, zero otherwise, and $\offset = \mathrm{median}_k
 f(\x_k)$ equals, for example, to the median or $25\%$-tile or best function 
 value of the feasible points
 in the same generation. If no other information is available, $c_i(\x)$ might be computed as the squared distance of $\x$ to the best or the closest feasible solution in the population or the closest known feasible solution. The latter is reminiscent to the boundary repair above. 
 This approach has not yet been experimentally
 evaluated by the author. A different, slightly more involved approach is given in \cite{collange2010multidisciplinary}.
 Similar and more recent approaches \cite{dufosse2021augmented} have already found there way into the \href{https://github.com/CMA-ES/pycma}{Python \texttt{cma} package}.

\end{description}

  In either case of \eqref{eq:penbound} and \eqref{eq:pencon}, $\alpha$
  should be chosen such that the differences in $f$ and the
  differences in the second summand have a similar magnitude.
  
\newpage
\section{\textsc{MATLAB} Source Code\label{sec:source}}
This code does not implement negative weights, that is, $w_i=0$ for $i>\mu$ in Table\nref{tabdefpara}. 
\begin{scriptsize}
\begin{verbatim}
  1 function xmin=purecmaes
  2   % CMA-ES: Evolution Strategy with Covariance Matrix Adaptation for
  3   % nonlinear function minimization. 
  4   %
  5   % This code is an excerpt from cmaes.m and implements the key parts
  6   % of the algorithm. It is intendend to be used for READING and
  7   % UNDERSTANDING the basic flow and all details of the CMA *algorithm*. 
  8   % Computational efficiency is sometimes disregarded. 
  9 
 10   % --------------------  Initialization --------------------------------  
 11 
 12   % User defined input parameters (need to be edited)
 13   strfitnessfct = 'felli'; % name of objective/fitness function
 14   N = 10;                  % number of objective variables/problem dimension
 15   xmean = rand(N,1);       % objective variables initial point
 16   sigma = 0.5;             % coordinate wise standard deviation (step-size)
 17   stopfitness = 1e-10;  % stop if fitness < stopfitness (minimization)
 18   stopeval = 1e3*N^2;   % stop after stopeval number of function evaluations
 19   
 20   % Strategy parameter setting: Selection  
 21   lambda = 4+floor(3*log(N));  % population size, offspring number
 22   mu = lambda/2;   % lambda=12; mu=3; weights = ones(mu,1); would be (3_I,12)-ES
 23   weights = log(mu+1/2)-log(1:mu)';     % muXone recombination weights
 24   mu = floor(mu);      % number of parents/points for recombination
 25   weights = weights/sum(weights);       % normalize recombination weights array
 26   mueff=sum(weights)^2/sum(weights.^2); % variance-effective size of mu
 27 
 28   % Strategy parameter setting: Adaptation
 29   cc = (4+mueff/N) / (N+4 + 2*mueff/N);  % time constant for cumulation for C
 30   cs = (mueff+2)/(N+mueff+5);  % t-const for cumulation for sigma control
 31   c1 = 2 / ((N+1.3)^2+mueff);  % learning rate for rank-one update of C and
 32   cmu = min(1-c1, 2*(mueff-2+1/mueff) / ((N+2)^2+2*mueff/2));  % for rank-mu update
 33   damps = 1 + 2*max(0, sqrt((mueff-1)/(N+1))-1) + cs; % damping for sigma 
 34   
 36   % Initialize dynamic (internal) strategy parameters and constants
 37   pc = zeros(N,1); ps = zeros(N,1);   % evolution paths for C and sigma
 38   B = eye(N);                         % B defines the coordinate system
 39   D = eye(N);                         % diagonal matrix D defines the scaling
 40   C = B*D*(B*D)';                     % covariance matrix
 41   eigeneval = 0;                      % B and D updated at counteval == 0
 42   chiN=N^0.5*(1-1/(4*N)+1/(21*N^2));  % expectation of 
 43                                       %   ||N(0,I)|| == norm(randn(N,1))
 44   
 45   % -------------------- Generation Loop --------------------------------
 46 
 47   counteval = 0;  % the next 40 lines contain the 20 lines of interesting code 
 48   while counteval < stopeval
 49
\end{verbatim}~\\[-5.33ex]
\tabverb| 50     
\tabverb| 51     for k=1:lambda,                                                      |\\
\tabverb| 52       arz(:,k) = randn(N,1);  
\tabverb| 53       arx(:,k) = xmean + sigma * (B*D * arz(:,k));   
\tabverb| 54       arfitness(k) = feval(strfitnessfct, arx(:,k)); 
\tabverb| 55       counteval = counteval+1;                                           |\\
\tabverb| 56     end                                                                  |\\
\tabverb| 57                                                                          |\\
\tabverb| 58     
\tabverb| 59     [arfitness, arindex] = sort(arfitness); 
\tabverb| 60     xmean = arx(:,arindex(1:mu))*weights;   
\tabverb| 61     zmean = arz(:,arindex(1:mu))*weights;   
\tabverb| 62                                                                          |\\
\tabverb| 63     
\tabverb| 64     ps = (1-cs)*ps + (sqrt(cs*(2-cs)*mueff)) * (B * zmean);            
\tabverb| 65     hsig = norm(ps)/sqrt(1-(1-cs)^(2*counteval/lambda))/chiN < 1.4+2/(N+1);  |\\
\tabverb| 66     pc = (1-cc)*pc + hsig * sqrt(cc*(2-cc)*mueff) * (B*D*zmean);       
\tabverb| 67                                                                          |\\
\tabverb| 68     
\tabverb| 69     C = (1-c1-cmu) * C ...                 
\tabverb| 70          + c1 * (pc*pc' ...                
\tabverb| 71                  + (1-hsig) * cc*(2-cc) * C) ...  
\tabverb| 72          + cmu ...                         
\tabverb| 73            * (B*D*arz(:,arindex(1:mu))) ...                              |\\
\tabverb| 74            *  diag(weights) * (B*D*arz(:,arindex(1:mu)))';               |\\
\tabverb| 75                                                                          |\\
\tabverb| 76     
\tabverb| 77     sigma = sigma * exp((cs/damps)*(norm(ps)/chiN - 1));               
\tabverb| 78 |\\[-5.33ex] 
\begin{verbatim}
 79     % Update B and D from C
 80     if counteval - eigeneval > lambda/(cone+cmu)/N/10  % to achieve O(N^2)
 81       eigeneval = counteval;
 82       C=triu(C)+triu(C,1)'; % enforce symmetry
 83       [B,D] = eig(C);       % eigen decomposition, B==normalized eigenvectors
 84       D = diag(sqrt(diag(D))); % D contains standard deviations now
 85     end
 86 
 87     % Break, if fitness is good enough
 88     if arfitness(1) <= stopfitness 
 89       break;
 90     end
 91 
 92     % Escape flat fitness, or better terminate? 
 93     if arfitness(1) == arfitness(ceil(0.7*lambda))
 94       sigma = sigma * exp(0.2+cs/damps); 
 95       disp('warning: flat fitness, consider reformulating the objective');
 96     end
 97     
 98     disp([num2str(counteval) ': ' num2str(arfitness(1))]);
 99 
100   end % while, end generation loop
101 
102   % -------------------- Final Message ---------------------------------
103 
104   disp([num2str(counteval) ': ' num2str(arfitness(1))]);
105   xmin = arx(:, arindex(1)); % Return best point of last generation.
106                              % Notice that xmean is expected to be even
107                              % better.
108   
109 % ---------------------------------------------------------------  
110 function f=felli(x)
111   N = size(x,1); if N < 2 error('dimension must be greater one'); end
112   f=1e6.^((0:N-1)/(N-1)) * x.^2;  % condition number 1e6
 \end{verbatim}
 \kom{
   if arfitness(1) == arfitness(min(1+floor(lambda/2), 2+ceil(lambda/4)))
       sigma = sigma * exp(0.2+cs/damps); 
   end
 }
\end{scriptsize}

\newpage
\section{Reformulation of Learning Parameter $\ccov$}  


For sake of consistency and clarity, we have reformulated the learning coefficients in \eqref{algcov} and replaced
\begin{eqnarray}
 \frac{\ccov}{\mucov} &\mathrm{with}& \cone\, \\
 \hspace{-2em}\ccov\rklam{1-\frac{1}{\mucov}} &\mathrm{with}& \cmu\quad\text{and} \\
 1 - \ccov &\mathrm{with}& 1 - \cone - \cmu 
\enspace,
\end{eqnarray}
and chosen (in \eqsref{eq-def-cone} and \eqrefadd{eq-def-cmu}) 
\newcommand{\alphacmu}{\ensuremath{\alpha_\mu^0}} 
\begin{eqnarray}
  \cone &=& \frac{2}{(n+1.3)^2+\mucov} \label{eq:newfirst}\\
  \cmu &=& \min\left(2\,\frac{\mucov-2\, + 
     \frac{1}{\mucov}}{(n+2)^2 +\mucov}\,,\, 1-\cone\right) 
\enspace,
\end{eqnarray}
 The resulting coefficients are quite similar to the previous. In
 contrast to the previous formulation, \cone\ becomes monotonic in
 $\mueff^{-1}$ and $\cone+\cmu$ becomes virtually monotonic in
 $\mueff$. 

\begin{figure}[tb]
 \vspace*{-1cm}
  \begin{center}
      \includegraphics[width=0.85\textwidth]{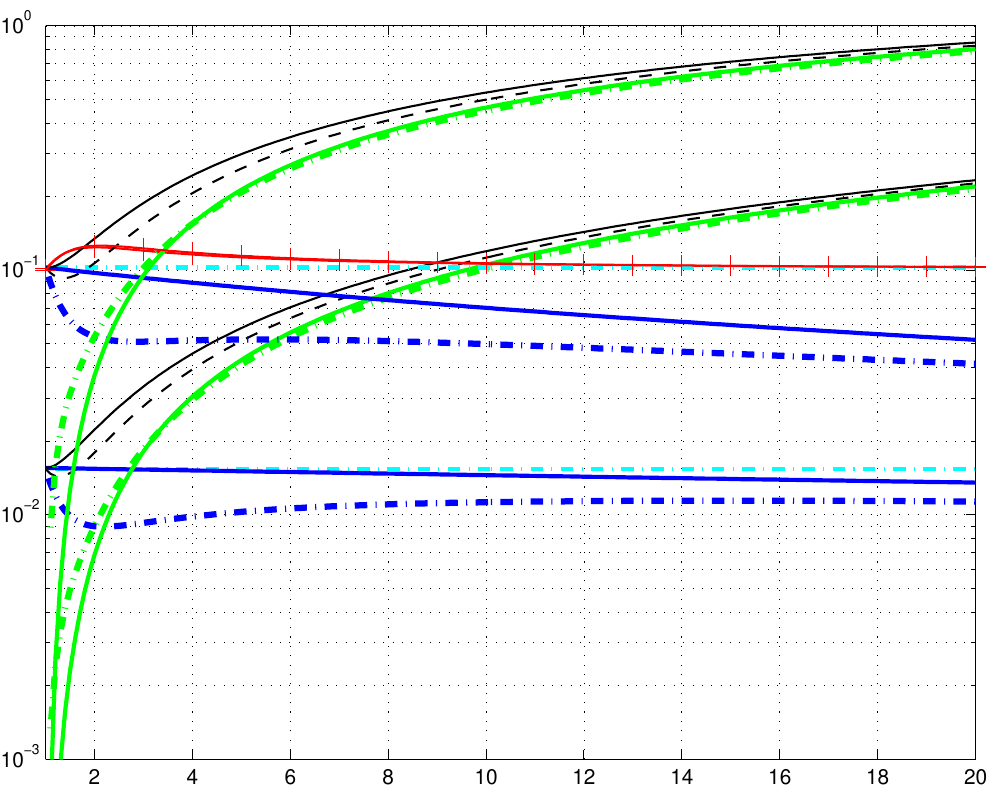}\\
      \includegraphics[width=0.85\textwidth]{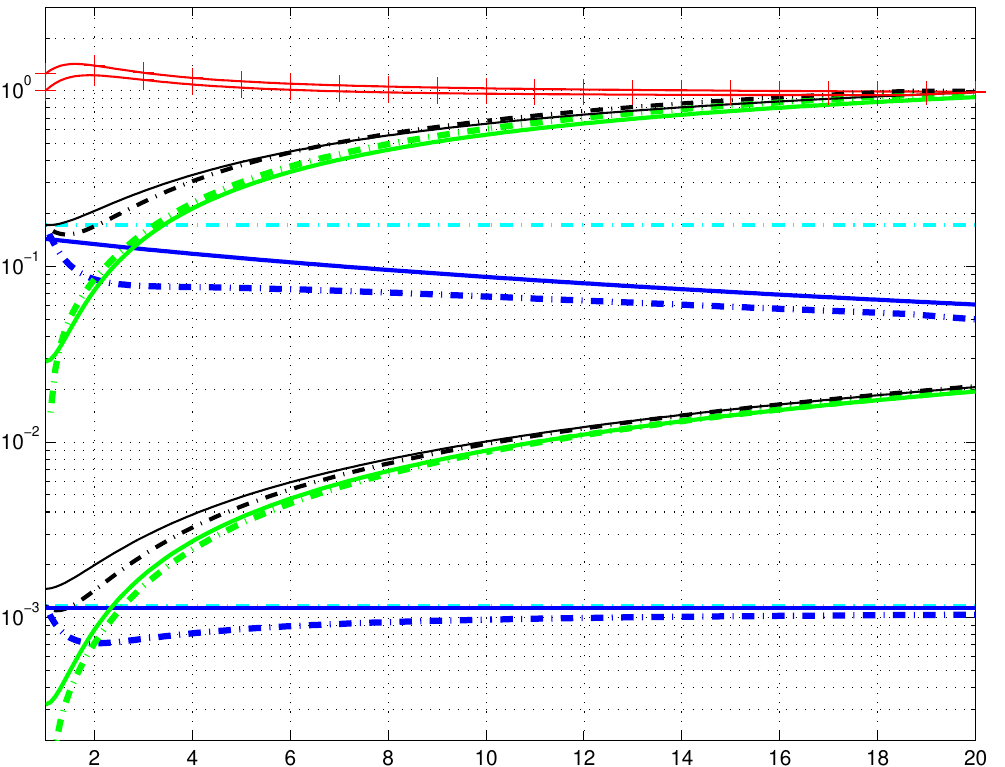}
 \vspace*{-0.3cm}
  \end{center}
  \caption[ccov]{\label{figccov} Learning rates $\cone,\cmu$ (solid)
  and \ccov\ (dash-dotted) versus $\mueff$.  Above: Equations
  \eqref{eq:newfirst} etc.\ for $n=3;10$. Below: Equations
  \eqref{eq:newalter} etc.\ for $n=2;40$. Black: $\cone+\cmu$ and
  $\ccov$; blue: \cone\ and $\ccov/\mucov$; green: \cmu\ and
  $(1-1/\mucov)\ccov$; cyan: $2/(n^2+\sqrt{2})$; red:
  $(\cone+\cmu)/\ccov$, above divided by ten. For $\mucov\approx2$ the
  difference is maximal, because $\cone$ decreases much slower with
  increasing $\mucov$ and $\ccov$ is non-monotonic in $\mucov$ (a main
  reason for the new formulation). }
\end{figure}

Another alternative, depending only on the degrees of freedom in the covariance matrix and additionally correcting for very small $\lambda$, reads 
\begin{eqnarray}
  \cone &=& \frac{\min(1,\lambda/6)}{m+2\sqrt{m}+\frac{\mueff}{n}} \label{eq:newalter}\\
  \cmu &=& \min\left(1-\cone\,,\,\frac{\alphacmu+\mueff-2\, +
     \frac{1}{\mueff}}{m +4\sqrt{m}+\frac{\mueff}{2}}\right)  \\
  \alphacmu &=& 0.3
\enspace,
\end{eqnarray}
where $m=\frac{n^2+n}{2}$ is the degrees of freedom in the covariance
matrix.  For $\mueff=1$, the coefficient \cmu\ is now chosen to be
larger than zero, as $\alphacmu>0$. Figure\nref{figccov} compares the
new learning rates with the old ones.

\end{appendix}

\end{document}